\documentclass[lettersize,journal]{IEEEtran}
\usepackage{amsmath,amsfonts}
\usepackage{algorithmic}
\usepackage{algorithm}
\usepackage{array}
\usepackage[caption=false,font=normalsize,labelfont=sf,textfont=sf]{subfig}
\usepackage{textcomp}
\usepackage{stfloats}
\usepackage{url}
\usepackage{verbatim}
\usepackage{graphicx}
\usepackage{cite}
\usepackage{booktabs}
\usepackage{orcidlink}  
\usepackage{threeparttable}
\usepackage{tabularx}

\newtheorem{definition}{Definition}

\newtheorem{proposition}{Proposition}
\newtheorem{lemma}{Lemma}
\newtheorem{corollary}{Corollary}

\begin{document}
\bstctlcite{IEEEexample:BSTcontrol}

\title{Probability Distribution Learning Framework and Its Application in Understanding Deep Learning}

\author{Binchuan Qi$^{\orcidlink{0000-0001-5832-1884}}$,~\IEEEmembership{},Wei Gong$^{\orcidlink{0000-0003-4030-4968}}$,~\IEEEmembership{Member,~IEEE,}, Li Li$^{\orcidlink{0000-0000-0000-0000}}$,~\IEEEmembership{Member,~IEEE,}
\thanks{This paper was produced by the IEEE Publication Technology Group. They are in Piscataway, NJ.}
\thanks{Manuscript received April 19, 2021; revised August 16, 2021.}}

\markboth{Journal of \LaTeX\ Class Files,~Vol.~14, No.~8, August~2021}%
{Shell \MakeLowercase{\textit{et al.}}: A Sample Article Using IEEEtran.cls for IEEE Journals}

\IEEEpubid{0000--0000/00\$00.00~\copyright~2021 IEEE}

\maketitle

\begin{abstract}
Despite its empirical success, deep learning still lacks a comprehensive theoretical understanding of model fitting and generalization. This paper proposes the probability distribution (PD) learning framework to analyze the optimization and generalization mechanisms of deep learning. 
Within this framework, the conditional distribution of labels given features is the primary learning target, with the loss function, prior knowledge, and model properties explicitly characterized. Under these formulations, we establish theoretical guarantees on optimizability, even in non-convex settings, and derive generalization error bounds that provide meaningful explanations for practical performance.
Specifically, we first prove theoretically that the Fenchel-Young loss is the natural and necessary choice for solving PD learning problems, thereby justifying the generality of conclusions based on this loss.  
Second, to capture the characteristics of deep neural networks (DNNs), we introduce the notions of $\mathcal{H}(\psi)$-convexity and $\mathcal{H}(\Psi)$-smoothness, which generalize the classical concepts of strong convexity and Lipschitz smoothness. Based on them, we provide a theoretical explanation for the effectiveness of SGD in training DNNs.  
Finally, we derive model-independent bounds on the expected risk and generalization error for trained models, revealing the influence of the training set size, regularization term, the mutual information between labels and features, and the information loss caused by model irreversibility on risk and generalization.
Based on our theoretical analysis and experimental validation, we believe that the PD learning framework facilitates a deeper and more unified theoretical understanding of deep learning.
\end{abstract}


\section{Introduction}
\label{sec:intro}
\IEEEPARstart{W}{hile} deep learning has achieved unprecedented success across various domains, several foundational questions remain unresolved within the classical learning theory framework. Key among these are: how do first-order optimization methods such as stochastic gradient descent (SGD) \cite{Ghadimi2013StochasticFA} succeed in optimizing highly non-convex, non-smooth, and often ill-conditioned objectives? How does the number of parameters influence model capacity and generalization? 

Classical optimization and generalization theories are insufficient to fully characterize the fitting and generalization behavior of deep neural networks (DNNs), calling for a deeper understanding of the interplay between data structure and model dynamics.
Despite the inherent complexity of deep learning optimization landscapes, empirical evidence consistently shows that gradient-based algorithms are remarkably effective at finding high-quality solutions in practice~\cite{He2015DeepRL, Zhang2016UnderstandingDL, Zhang2021UnderstandingDL}.  
From the perspective of classical optimization theory, however, these algorithms are generally expected to converge only to stationary points~\cite{Lee2016GradientDO, Jentzen2018StrongEA}.  Furthermore, even seemingly simple non-convex problems, such as standard quadratic programming and copositivity testing, are known to be NP-hard in general~\cite{murty1985some,bellare1995complexity,pardalos1991quadratic,nesterov2008advance}.  
However, recent work by Qi et al.~\cite{Qi2025} provides new insight by modeling the supervised classification problem in deep learning as estimating the conditional distribution of labels given features. They theoretically show that, when using the Fenchel-Young loss, the squared norm of the loss gradient is equivalent to the mean squared error between the predicted and target conditional distributions, offering new insights into non-convex optimization dynamics.

In terms of generalization performance,  empirical observations reveal that high-capacity DNNs can achieve strong generalization performance on real data while failing to generalize at all when trained on randomly labeled datasets \cite{He2015DeepRL,Zhang2016UnderstandingDL,Zhang2021UnderstandingDL}. This behavior contradicts classical notions of overfitting based on VC theory \cite{vapnik1998statistical} and Rademacher complexity \cite{Bartlett2003RademacherAG}.
To address the limitations of classical theories in explaining the behavior of deep neural networks (DNNs), several modern theoretical frameworks have been proposed and actively developed. 
Among the most influential are theories derived from the perspective of overparameterization~\cite{Yun2018SmallNI, Arjevani2022AnnihilationOS}, such as the Neural Tangent Kernel (NTK)~\cite{Jacot2018NeuralTK, wang2023ntk, liu2024ntk} and mean-field theory~\cite{Sirignano2018MeanFA, Mei2018AMF, kim2024transformers}. However, for finite-width networks, it remains unclear whether these theories fully capture empirical convergence behavior~\cite{Seleznova2020AnalyzingFN,Vyas2023EmpiricalLO}.
Another prominent hypothesis attributes the success of stochastic gradient descent (SGD) to its implicit regularization, which tends to guide optimization toward flat minima believed to generalize better~\cite{Dauphin2014IdentifyingAA, Keskar2016OnLT, Lee2019FirstorderMA, zhou2023class, ding2024flat, le2024gradient, zou2024flatten}. Yet, recent results show that flatness measures can be manipulated without affecting generalization~\cite{dinh2017sharp}, suggesting that flat minima alone may not be a universal predictor of generalization performance.
Furthermore, the information bottleneck principle~\cite{ShwartzZiv2017OpeningTB, Ahuja2021InvariancePM, Wongso2023UsingSM} interprets learning as an information compression process in neural networks, where models aim to retain only the most relevant information about the target variable while discarding redundant details. However, Saxe et al.~\cite{saxe2019information} argue that such compression is not consistently observed in many neural networks, and that networks without compression, such as invertible ones, can still generalize well. Moreover, information-theoretic approaches often model parameters as random variables, which may lead to challenges in accurately estimating mutual information, including cases where the mutual information is infinite or intractable. These issues limit the practical applicability of information-based methods in real-world neural network analysis.

Although numerous theories on the learning and generalization performance of deep learning have been proposed, elucidating the generalization mechanisms underlying deep learning remains an open problem at present~\cite{Oneto2023DoWR}.

This paper aims to advance the theoretical understanding of deep learning. Our main contributions are summarized as follows:
\begin{itemize}
\item We propose the PD learning framework, in which the conditional probability distribution of labels given features serves as the learning target. Within this framework, we explicitly specify the requirements that the loss function, model, and prior knowledge must satisfy, thereby establishing a principled foundation for analyzing deep learning dynamics.

\item We prove that, under the PD learning framework with a distribution-based objective, any valid loss function must be equivalent to the Fenchel-Young loss. This result allows us to focus on the Fenchel-Young loss for theoretical analysis without loss of generality, simplifying the study of optimization and generalization while preserving broad applicability.

\item To better characterize deep models and their training objectives, we generalize the classical concepts of strong convexity and Lipschitz smoothness to $\mathcal{H}(\psi)$-convexity and $\mathcal{H}(\psi)$-smoothness, respectively. Under these extended conditions, we establish theoretically and validate experimentally that gradient energy and the extreme eigenvalues of the structure matrix are two key factors governing convergence to the global optimum. This analysis provides a theoretical explanation for the effectiveness of SGD in training DNNs within the PD learning framework.

\item We derive model-independent upper and lower bounds on the expected risk, empirical risk, and generalization error. These bounds reveal the influence of the training set size, the regularization, the mutual information between labels and features, and the information loss caused by model irreversibility on risk and generalization.
\end{itemize}

\textbf{Organization}. The paper is structured as follows: Section~\ref{sec:related_work} reviews related work, and Section~\ref{sec:pre} introduces preliminary concepts and lemmas. The PD learning framework is formally defined in Section~\ref{sec:framework}. Sections~\ref{sec:loss} and~\ref{sec:optimization} analyze the theoretical properties of the Fenchel-Young loss and the non-convex optimization dynamics, respectively. Section~\ref{sec:bounds} derives bounds on generalization, expected, and empirical risks. Experimental settings and results are presented in Section~\ref{sec:experiments}, and conclusions are summarized in Section~\ref{sec:conclusion}. All proofs are provided in the appendix, Section~\ref{appendix:proof}.

\section{Related work}
\label{sec:related_work}

Currently, theoretical efforts to address the unresolved mysteries in deep learning have predominantly followed two main avenues: expressivity (fitting capacity) and generalization ability. 

Essentially, expressivity is characterized by the capability of DNNs to approximate any function. The universal approximation theorem~\cite{Hornik1989MultilayerFN,Cybenko1989ApproximationBS,Leshno1993OriginalCM} states that a feedforward network with any "squashing" activation function $\sigma(\cdot)$, such as the logistic sigmoid function, can approximate any Borel measurable function $f(x)$ with any desired non-zero amount of error, provided that the network is given enough hidden layer size $n$~\cite{Hanin2017ApproximatingCF,Kidger2019UniversalAW,Zhang2023GoingDG}. Recently, Qi et al.~\cite{Qi2025} framed supervised classification as learning the conditional label distribution and showed, both theoretically and empirically, that minimizing the Fenchel-Young loss $d_\Phi(\cdot,\cdot)$ reduces the MSE between predicted and empirical distributions. However, three key limitations remain: (1) it is unclear whether minimizing the gradient norm also minimizes the loss itself; (2) the reliance on Fenchel-Young loss lacks generality; and (3) the generalization mechanism from a distribution learning perspective is unaddressed.  
Building on this, we formalize deep learning as conditional distribution learning and study both optimization and generalization within this unified framework.

Generalization ability represents a model's predictive ability on unseen data. Increasing the depth of a neural network architecture naturally yields a highly overparameterized model, whose loss landscape is known for a proliferation of local minima and saddle points~\cite{Dauphin2014IdentifyingAA}. The fact that sometimes (for some specific models and datasets), the generalization ability of a model paradoxically increases with the number of parameters, even surpassing the interpolation threshold, where the model fits the training data perfectly. This observation is highly counterintuitive since this regimen usually corresponds to over-fitting. DNNs show empirically that after the threshold, sometimes the generalization error tends to descend again, showing the so-called Double-Descent generalization curve, and reaching a better minimum. To address this question, researchers have tried to focus on different methods and studied the phenomena under different names, e.g., over-parametrization~\cite{Du2018GradientDP,Arjevani2022AnnihilationOS}, double descent~\cite{Poggio2019DoubleDI,Liu2021OnTD}, sharp/flat minima~\cite{Hochreiter1997FlatM,dinh2017sharp,Kaddour2022WhenDF,Wu2022TheAP}, and benign overfit~\cite{Koehler2021UniformCO,Chen2022UnderstandingBO}. We review the related literature and organize it into the following six categories:
\textit{Complexity-based methods.}~\cite{Vapnik2006EstimationOD,Bartlett2003RademacherAG,Wu2021StatisticalLT}.  
\textit{Algorithmic stability.}~\cite{Bousquet2002StabilityAG,Oneto2015FullyEA,ANDREASMAURER2017ASL}.  
\textit{PAC-Bayesian methods.}~\cite{ShaweTaylor1997APA,McAllester1998SomePT,Xie2017PACBayesBF,Alquier2021UserfriendlyIT}.  
\textit{Implicit regularization.}~\cite{Bottou2016OptimizationMF,Jastrzebski2017ThreeFI,Lewkowycz2020TheLL}.  
\textit{Over-parameterization.}~\cite{Jacot2018NeuralTK,Du2018GradientDP,Mei2018AMF,Chizat2018OnTG}.  
\textit{Information-based methods.}~\cite{Xu2017InformationtheoreticAO,Hellstrm2022ANF,Wongso2023UsingSM}. Despite extensive research, there remains no clear understanding of why, for certain algorithms and datasets, increasing the number of parameters can actually improve generalization~\cite{Oneto2023DoWR}. 
Nevertheless, accumulating evidence and existing theories suggest that the remarkable performance of deep neural networks arises from a combination of factors, including model architecture, optimization dynamics, and properties of the training data.  
Elucidating the underlying mechanisms of deep learning remains an open challenge.

\section{Preliminaries}
\label{sec:pre}

\subsection{Notation}

    \par \textit{1. }Random variables are denoted using upper case letters such as $Z$, $X$, and $Y$, which take values in sets $\mathcal{Z}$, $\mathcal{X}$, and $\mathcal{Y}$, respectively. The cardinality of a set $\mathcal{Z}$ is denoted by $|\mathcal{Z}|$. 
    \par \textit{2. }We utilize the notation $f_\theta$ (hereinafter abbreviated as $f$) to denote the model characterized by the parameter vector $\theta$. The space of models, which is a set of functions endowed with some structure, is represented by $\mathcal{F}_{\Theta}=\{f_\theta:\theta\in \Theta\}$, where $\Theta$ denotes the parameter space. 
    \par \textit{3. }$I_C$ is the indicator function ($I_C(q) = 0$ if $q \in C$, else $+\infty$)
    \par \textit{4. }The Legendre-Fenchel conjugate of a function $\Phi$ is denoted by $\Phi^*(\nu):= \sup_{\mu \in \mathrm{dom}(\Phi)}\langle \mu,\
    \nu \rangle-\Phi(\mu)$~\cite{Todd2003ConvexAA}. By default, $\Phi$ is a continuous strictly convex function, and its gradient with respect to $\mu$ is denoted by $\nabla_\mu \Phi(\mu)$. For convenience, we use $\mu_\Phi^*$ to represent $\nabla_\mu \Phi(\mu)$. When $\Phi(\cdot)=\frac{1}{2}\|\cdot\|_2^2$, we have $\mu=\mu _\Phi^*$. The Fenchel-Young loss $d_\Phi \colon \mathrm{dom}(\Phi) \times \mathrm{dom}(\Phi^*) \to \mathbb{R}_{\ge 0}$ \label{def:FY_loss} generated by $\Phi$ is defined as:
\begin{equation}
d_{\Phi}(\mu, \nu) 
:= \Phi(\mu) + \Phi^*(\nu) - \langle \mu,\nu\rangle,
\label{eq:fy_losses}
\end{equation}
where $\Phi^*$ denotes the conjugate of $\Phi$.
\par \textit{5. }The maximum and minimum eigenvalues of a matrix $A$ are denoted by $\lambda_{\max}(A)$ and $\lambda_{\min}(A)$, respectively.
\par  \textit{6. }To improve clarity and conciseness, we transform the distribution function into a vector for processing and provide the following definitions of symbols:
\begin{equation}
    \begin{aligned}
        &q_{\mathcal{X}}:=(q_{X}(x_{1}),\cdots, q_{X}(x_{|\mathcal X|}))^\top,\\
        &q_{\mathcal{Y}|x}:=(q_{Y|X}(y_{1}|x),\cdots,q_{Y|X}(y_{|\mathcal Y|}|x))^\top.
    \end{aligned}
\end{equation}
Here, $q_{X}(x)$ and $q_{Y|X}(y|x)$ represent the marginal and conditional PMFs/PDFs, respectively. 
\par \textit{7. }We define the generalized entropy of the conditional distribution as
\begin{equation}
    \mathrm{Ent}_\Phi(q_{\mathcal{Y}|X}) = \mathbb{E}_X[\Phi(q_{\mathcal{Y}|X})] - \Phi(q_{\mathcal{Y}}),
\end{equation}
~\cite{10.1093/acprof:oso/9780199535255.001.0001}where $\Phi$ is a convex function, ensuring that the generalized entropy is non-negative.  Let $\mathbf{1}_y$ denotes the indicator vector (one-hot encoding) for label $y$, with dimension $|\mathcal{Y}|$, where the component corresponding to class $y$ is 1 and all others are 0. Based on the definition of generalized entropy, we further define
\begin{equation}
    \begin{aligned}
        \mathrm{Ent}_\Phi(\mathbf{1}_Y|x) &= \mathbb{E}_{Y\sim q_{\mathcal{Y}|x}}\Phi(\mathrm{1}_Y) - \Phi(q_{\mathcal{Y}|x}), \\
        \mathrm{Ent}_\Phi(\mathbf{1}_Y|X) &= \mathbb{E}_{X\sim q_{\mathcal{X}}} \left[ \mathbb{E}_{Y\sim q_{\mathcal{Y}|x}}\Phi(\mathrm{1}_Y) - \Phi(q_{\mathcal{Y}|x}) \right].
    \end{aligned}
\end{equation}

When $\Phi(q) = \sum_i q_i \log q_i$, we have $\mathrm{Ent}_\Phi(q_{\mathcal{Y}|X}) = I(Y;X)$, i.e., the Shannon mutual information, and $\mathrm{Ent}_\Phi(\mathbf{1}_Y|X) = H(Y|X)$, which corresponds to the Shannon conditional entropy. Therefore, we refer to $\mathrm{Ent}_\Phi(q_{\mathcal{Y}|X})$ and $\mathrm{Ent}_\Phi(\mathbf{1}_Y|X)$ as the generalized mutual information and generalized conditional entropy, respectively\label{def:generalized_entropy}.

\subsection{Lemmas}
The theorems and propositions in this paper rely on the following lemmas. 

\begin{lemma}
\label{lem:sum_conjugate}
It is well-known that when $\Phi = \Psi + I_C$, it follows that $\Phi^*(x) = \inf_{x' \in \mathbb R^d} \sigma_C(x') + \Psi^*(x -
x')$, where we used $I_C^*(x) = \sigma_C(x)= \max_{y
\in C} \langle x,y\}$ \cite{Todd2003ConvexAA,Beck2012SmoothingAF} . 
\end{lemma}

\begin{lemma}[Euler's theorem for homogeneous functions]
\label{lem:euler}
If $f: \mathbb{R}^n \to \mathbb{R}$ is a differentiable $k$-homogeneous function, then $
x \cdot \nabla f(x) = k f(x)$, 
where $x \cdot \nabla f(x) = \sum_{i=1}^n x_i \frac{\partial f}{\partial x_i}$.
\end{lemma}

\begin{lemma}
    \label{lem:l1_convergence}
Let the random variable $X \sim q$ have a distribution supported on $k$ points (i.e., the cardinality of its sample space is $k$), and let $\hat{q}^n$ denote the empirical distribution of $X$ constructed from $n$ independent and identically distributed (i.i.d.) samples. Then, for any $\varepsilon \ge \sqrt{\frac{20k}{n}}$, the following inequality holds~\cite[Lemma 3]{10.1214/aos/1176346255}: $
    \Pr\left(\|q - \hat{q}^n\|_1 \ge \varepsilon\right) \le 3\exp\left(-\frac{n\varepsilon^2}{25}\right)$.
\end{lemma}

\subsection{Setting and basics}
\label{subsec:basic_setting}
Let $Z = (X,Y)$ be a random pair drawn from the unknown true underlying distribution $q$, where $(X,Y) \in \mathcal{Z} = \mathcal{X} \times \mathcal{Y}$. Without loss of generality, we assume that both $\mathcal{X}$ and $\mathcal{Y}$ are finite-dimensional spaces.
Given a training dataset $s^n := \{z^{(i)}\}_{i=1}^n = \{(x^{(i)}, y^{(i)})\}_{i=1}^n$, consisting of $n$ i.i.d. samples drawn from $q_{\mathcal{Z}}$ (abbreviated as $q$).
The empirical probability distribution based on the sample $s^n$ is defined as $
\hat{q}(s^n) := \frac{1}{n} \sum_{i=1}^n \mathbf{1}_{\{z\}}(z^{(i)})$, which we refer to as the empirical distribution.

\section{Framework of PD learning}
\label{sec:framework}

In this section, we introduce and formally define the PD learning framework. 

\subsection{Problem formulation and framework definition}
Recently, Qi et al.~\cite{Qi2025} modeled the supervised classification problem in deep learning as learning the conditional probability distribution of labels given features using the Fenchel-Young loss. They proved that under this formulation, SGD can successfully optimize the non-convex training objective of DNNs, and they provided theoretical explanations for the roles of random initialization, over-parameterization, and skip connections in optimization. Motivated by these insights, we now formally introduce the PD learning framework.

\begin{definition}[PD learning]
    \label{def:pd_learning_problem}
Given i.i.d. samples $s^n$, prior knowledge $C$ about the conditional distribution $q_{\mathcal{Y}|x}$, a model space $\mathcal{F}_\Theta$, and a strictly convex function $\Phi$, solve the following optimization problem:
$$
\min_{\theta \in \Theta} \mathbb{E}_{X,Y} \left[ \ell(\mathbf{1}_Y, f_\theta(X)_{\Phi^*}^*) \right],
$$
subject to the following requirements on the loss function, prior knowledge, and model space: 
    \par \textit{Loss function}: The loss $\ell(\mu, \nu)$ must satisfy:
    \begin{itemize}
        \item \textit{Extended convexity}: $\ell(\mu, \nu)$ is $\mathcal{H}(\psi)$-convex in $\nu$ (see Definition~\ref{def:g_convex_smooth}).
        \item \textit{Expectation optimality}: The expected loss $\mathbb{E}_{\mu}[\ell(\mu, \nu)]$ achieves its global minimum at $\nu = \mathbb{E}_\mu[\mu]$.
    \end{itemize}
    \par \textit{Model}: The composition $\ell(\mu, f_\theta(x)_{\Phi^*}^*)$ is $\mathcal{H}(\xi)$-smooth with respect to the parameters $\theta$ (see Definition~\ref{def:g_convex_smooth}).
    \par \textit{Prior Knowledge}: $C$ is a convex set consisting of all $q_{\mathcal{Y}|x}$ satisfying given constraints. In the absence of prior knowledge, $C$ defaults to the probability simplex, i.e., $C = \Delta$.
\end{definition}

Next, we briefly explain the rationale and physical interpretation behind the optimization objective of PD learning and the imposed constraints on the loss function, prior knowledge and model. Detailed discussions and justifications for the ubiquity of these conditions in practical deep learning tasks will be provided in Sections~\ref{sec:loss} and \ref{sec:optimization}.

    \par In the PD learning framework, the loss function is required to be \textit{extended convex}, meaning $\ell(\mu, \nu)$ is $\mathcal{H}(\psi)$-convex in $\nu$. By the definition of $\mathcal{H}(\psi)$-convexity, this implies that the loss is strictly convex in $\nu$, ensuring a unique minimum at $\mu = \nu$. This property is essential to guarantee that the optimization process converges to a unique optimal solution. Moreover, $\mathcal{H}(\psi)$-convexity generalizes strong convexity and applies to a broader class of functions, enhancing the framework's adaptability across different models and tasks.

    \par PD learning also requires \textit{expectation optimality} for the loss function: the expected loss $\mathbb{E}_{\mu}[\ell(\mu, \nu)]$ must attain its global minimum at $\nu = \mathbb{E}_\mu[\mu]$. This ensures the model learns the true conditional distribution of labels given features. Specifically, when labels are one-hot encoded, $\mathbb{E}_{Y|x}[\ell(\mathbf{1}_Y, \nu)]$ is minimized at $\nu = \mathbb{E}_{Y|x}[\mathbf{1}_Y] = q_{\mathcal{Y}|x}$, so the optimal prediction matches the conditional distribution.

    \par PD learning requires the composite objective $\ell(\mu, f_\theta(x)_{\Phi^*}^*)$ to be $\mathcal{H}(\xi)$-smooth with respect to $\theta$. This property ensures that the SGD algorithm can effectively minimize the gradient energy~\ref{def:gradient_energy}. The theoretical mechanisms and proofs will be elaborated in \ref{sec:optimization}.

    \par The PD learning framework incorporates prior knowledge $C$, which is leveraged to design the loss function in a way that reduces optimization difficulty and improves efficiency. For any differentiable strictly convex function $\Omega$, we define the loss function as $\Phi(\cdot) = \Omega(\cdot) + I_C(\cdot)$. This construction ensures that $\mathrm{dom}(\Phi) \subseteq C$. When $\Phi^*(f_\theta(x))$ is finite and differentiable, Legendre-Fenchel duality implies that the associated primal point $f_\theta(x)_{\Phi^*}^* \in \mathrm{dom}(\Phi) \subseteq C$. Consequently, the model's prediction, interpreted as $f_\theta(x)_{\Phi^*}^*$, is naturally constrained within the set $C$ without requiring explicit projection or regularization. 

In Section~\ref{sec:loss}, we prove that any  loss function satisfying extended convexity and expectation optimality is equivalent to the Fenchel-Young loss $d_\Phi(\mu, \nu_{\Phi}^*)$. Based on this equivalence, the optimization objective of PD learning can be uniformly expressed as:
\begin{equation}
\min_{\theta} \mathbb{E}_{(X,Y) \sim q} \left[ d_\Phi(\mathbf{1}_Y, f_\theta(X)_{\Phi^*}^*) \right],
\end{equation}
where $f_\theta(x)_{\Phi^*}^* = \nabla \Phi^*(f_\theta(x))$, and $\Phi^*$ denotes the convex conjugate of a strictly convex function $\Phi$. This formulation explicitly links the loss function to the structure of convex conjugacy, providing a unified and geometrically meaningful framework for theoretical analysis. Therefore, all subsequent theoretical developments in this work are based on the Fenchel-Young loss representation.

\begin{figure}[ht]
\centering
\includegraphics[width=\columnwidth]{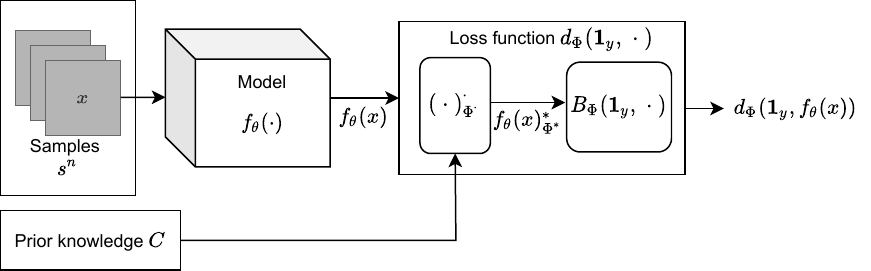}
\caption{Schematic illustration of the PD learning framework. This diagram presents the complete information processing pipeline from input data to latent distribution approximation, highlighting the relationships among model prediction, conjugate transformation, and distributional distance measurement.}
\label{fig:pd_structure}
\end{figure}

To intuitively illustrate the overall architecture and information flow of PD learning, Figure~\ref{fig:pd_structure} depicts the data processing pipeline. Starting from the raw input $x$, the framework maps it through a parametric model $f_\theta$ to produce a prediction, which is then transformed via the convex conjugate gradient $(\cdot)_{\Phi^*}^*$ into a distributional representation. Then, the Bregman divergence is computed between this representation and the one-hot encoded ground truth $\mathbf{1}_y$, forming the optimization objective $d_\Phi(\mathbf{1}_y,f_\theta(x))$. 

In essence, PD learning can be viewed as a structured instantiation of the classical statistical learning framework. While it retains the fundamental components, sampled data (training examples), a loss function, and a model (hypothesis) space, it introduces an additional, crucial element: prior knowledge $C$. This prior knowledge is encoded as a convex set of admissible conditional distributions $q_{\mathcal{Y}|x}$, allowing the framework to incorporate domain-specific constraints directly into the learning process. 
We will demonstrate in the following sections that, by integrating prior knowledge and imposing structured constraints on the loss function and model, the PD learning framework achieves superior analytical tractability and provides strong explanatory power for existing deep learning problems. 

\subsection{Definitions of key terms in PD learning }
To clearly articulate our approach, we present the following definitions used within the PD learning framework:

    \par \textit{Expected and Empirical Risk}. In Section~\ref{sec:loss}, we prove that under general conditions, any standard loss function satisfying extended convexity and expectation optimality is equivalent to the Fenchel-Young loss $d_\Phi(\mu, \nu_{\Phi}^*)$. Using this standard loss function $d_\Phi(\cdot,\cdot)$, the expected risk and empirical risk in PD learning are defined, respectively, as:
    \begin{equation}
    \begin{aligned}
           \quad \mathcal{L}_{\Phi}(q,f_\theta(\cdot))  &=  \mathbb{E}_{(X,Y)\sim q}\, d_{\Phi}(\mathbf{1}_Y,f_\theta(X)),\\
            \quad \mathcal{L}_{\Phi}(\hat{q},f_\theta(\cdot)) &=  \mathbb{E}_{(X,Y)\sim \hat{q}}\, d_{\Phi}(\mathbf{1}_Y,f_\theta(X)),
    \end{aligned}
    \end{equation}
    where $\mathbf{1}_y$ denotes the indicator vector (one-hot encoding) for label $y$. In the special case where $\Phi(\cdot) = \frac{1}{2}\|\cdot\|_2^2$, the expected risk for PD learning becomes:
    \begin{equation}
    \begin{aligned}
            \mathcal{L}_{L_2^2/2}(q,f_\theta(\cdot))  &=  \mathbb{E}_{(X,Y)\sim q}\|\mathbf{1}_{Y}-f_{\theta}(X)\|_2^2.
    \end{aligned}
    \end{equation}
    \par \textit{Distribution Fitting Error}. This metric evaluates the model's ability to fit the target distribution. Depending on the target distribution, we define the \textit{expected distribution fitting error} as:
    \begin{equation}
        \mathbb{E}_{X\sim q_{\mathcal{X}}} d_\Phi(q_{\mathcal{Y}|X},f_{\theta}(X)),
    \end{equation} 
    and the \textit{empirical distribution fitting error} as:
\begin{equation}
        \mathbb{E}_{X\sim \hat{q}_{\mathcal{X}}} d_\Phi(\hat{q}_{\mathcal{Y}|X},f_{\theta}(X)).
    \end{equation}
    In particular, when $\Phi(\cdot) = \frac{1}{2}\|\cdot\|_2^2$, the expected and empirical fitting errors become $\mathbb{E}_{X\sim q_{\mathcal{X}}} \|q_{\mathcal{Y}|X}-f_{\theta}(X)_{\Phi^*}^*\|_2^2$ and $\mathbb{E}_{X\sim \hat{q}_{\mathcal{X}}} \|\hat{q}_{\mathcal{Y}|X}-f_{\theta}(X)_{\Phi^*}^*\|_2^2$, respectively.
    \par\label{def:gradient_energy}\textit{Gradient Energy}. We define the gradient energy for an individual sample pair $(x,y)$ and for a dataset $s$ as:
\begin{equation}
    \begin{aligned}
        &\mathcal{G}_\Phi(\theta, (x,y)) := \|\nabla_\theta d_\Phi(\mathbf{1}_{y}, f_\theta(x))\|_2^r, \\
        &\mathcal{G}_\Phi(\theta, s) := \mathbb{E}_{(X,Y)\sim \hat{q}} \|\nabla_\theta d_\Phi(\mathbf{1}_{Y}, f_\theta(X))\|_2^r,
    \end{aligned}
\end{equation}
respectively.
    \par \textit{Structure Matrix}. We define the structure matrix associated with the model $f_\theta$ and input feature $x$ as:
    \begin{equation}
        A_x(\theta) := \nabla_\theta f_\theta(x)^\top \nabla_\theta f_\theta(x).
    \end{equation}

\section{Loss function and prior knowledge in PD learning}
\label{sec:loss}

\subsection{Unified form of the loss function}
It is found that a wide variety of loss functions commonly used in practical machine learning applications can be expressed in the form of Fenchel-Young losses~\cite{Blondel2019LearningWF}. In this section, we further establish the theoretical generality of Fenchel-Young losses by showing that any loss function satisfying the core conditions of PD learning, namely, extended convexity and expectation optimality, is equivalent to a Fenchel-Young loss. This equivalence is fundamental: it ensures that the structural form of the loss function within the PD learning framework is essentially unique. Consequently, our theoretical analysis based on Fenchel-Young losses is not only rigorously justified but also universally applicable across a broad spectrum of practical learning scenarios. 

The extended convexity condition implies that the loss function $\ell(q, p)$ is strictly convex with respect to $p$. The optimal expectation condition requires that the expected loss $\mathbb{E}_{Y\sim \hat{q}_{\mathcal{Y}|x}}[\ell(\mathbf{1}_Y, p)]$ achieves its global minimum when $p = \hat{q}_{\mathcal{Y}|x}$. This property ensures that minimizing the expected loss drives the model to learn the true conditional probability distribution of the labels given the input features. Together with the (Weak) Law of Large Numbers, these conditions guarantee that the empirical risk converges asymptotically to the expected risk as the sample size increases. Consequently, minimizing the empirical risk leads the model predictions to converge to the true underlying conditional probability distribution $q_{\mathcal{Y}|x}$, ensuring statistically consistent learning. We now present a key theoretical result:

\begin{proposition}[Uniqueness of Fenchel-Young Representation]
\label{prop:uniqueness_fenchel}
If a loss function $\ell(q, p)$ satisfies both extended convexity and optimal expectation, then there exists a strictly convex function $\Phi$ such that:
\begin{equation}
\ell(q, p) - \ell(q, q) = d_\Phi(q, p_\Phi^*),
\end{equation}
where $d_\Phi(\cdot, \cdot)$ denotes the Fenchel-Young loss induced by $\Phi$, and $p_\Phi^* = \nabla \Phi^*(p)$ represents the gradient of the convex conjugate of $\Phi$ evaluated at $p$.
\end{proposition}
The proof of this result is provided in Appendix~\ref{appendix:proof_uniqueness_fenchel}.

This proposition establishes that any loss function satisfying the above regularity conditions can be equivalently represented as a Fenchel-Young loss up to an additive constant (the loss on the diagonal, $\ell(q,q)$). As an illustrative example, consider the Cross-Entropy loss $\ell_{CE}(\cdot, \cdot)$ combined with the Softmax function. It can be reformulated as:
\begin{equation}
    \begin{aligned}
            \ell_{CE}(q_{\mathcal{Y}|x}, p_{\mathcal{Y}|x}) - \ell_{CE}(q_{\mathcal{Y}|x}, q_{\mathcal{Y}|x}) &= d_\Phi(q_{\mathcal{Y}|x}, f_\theta(x)),
    \end{aligned}
\end{equation}
where $\Phi(p_{\mathcal{Y}|x}) = -H(p_{\mathcal{Y}|x})$ is the negative entropy of $p_{\mathcal{Y}|x}$, and $p_{\mathcal{Y}|x} = (f_\theta(x))_{\Phi^*}^* = \nabla \Phi^*(f_\theta(x)) = \mathrm{Softmax}(f_\theta(x))$ is the probability distribution obtained by applying the softmax transformation to the model output $f_\theta(x)$. In this case, the Softmax Cross-Entropy loss emerges naturally as the Fenchel-Young loss associated with the negative entropy. In this formulation, the model’s predicted distribution is given by  
\begin{equation}
p_{\mathcal{Y}|x} = (f_\theta(x))_{\Phi^*}^* = \nabla \Phi^*(f_\theta(x)),
\end{equation}
which corresponds to applying the Legendre–Fenchel transform (i.e., the gradient of the conjugate function) to the model output $f_\theta(x)$ to obtain a valid probability distribution that approximates the target $q_{\mathcal{Y}|x}$. The Softmax function is a special case of $\nabla \Phi^*(\cdot)$, corresponding to the negative entropy generator. Therefore, all subsequent theoretical developments in this work are based on the Fenchel-Young loss representation, ensuring both mathematical rigor and broad applicability across various deep learning paradigms.

We decompose the expected and empirical risks into model-dependent and model-independent components, leading to the following proposition. 
\begin{proposition}[Decomposition of Risk in PD learning]
    \label{prop:decompose}
The expected and empirical risks can be decomposed as:\small
\begin{equation}
\begin{aligned}
\mathcal{L}_{\Phi}(q,f_\theta(\cdot)) &=  \mathrm{Ent}_\Phi(\mathbf{1}_Y|X)+ \mathbb{E}_{X} d_\Phi(q_{\mathcal{Y}|X}, f_\theta(X)), \\
\mathcal{L}_{\Phi}(\hat{q},f_\theta(\cdot)) &= \mathrm{Ent}_\Phi(\mathbf{1}_{\hat{Y}}|\hat{X}) + \mathbb{E}_{\hat{X}} d_\Phi(\hat{q}_{\mathcal{Y}|\hat{X}}, f_\theta(\hat{X})),
\end{aligned}
\end{equation}
where $(XY)\sim q$, $(\hat{X}\hat{Y})\sim \hat{q}$. 
\end{proposition}
The proof of this result is provided in Appendix~\ref{appendix:proof_decompose}.

Since $\mathrm{Ent}_\Phi(\mathbf{1}_Y|X)$ and $\mathrm{Ent}_\Phi(\mathbf{1}_{\hat{Y}}|\hat{X})$ are constants independent of the model $f_\theta$, minimizing the expected or empirical risk is equivalent to minimizing the distribution fitting errors $\mathbb{E}_{X\sim q_{\mathcal{X}}} d_\Phi(q_{\mathcal{Y}|X}, f_\theta(X))$ and $\mathbb{E}_{X\sim \hat{q}_{\mathcal{X}}} d_\Phi(\hat{q}_{\mathcal{Y}|X}, f_\theta(X))$, respectively. Proposition~\ref{prop:decompose} reveals that \textit{minimizing the expected and empirical risks is fundamentally equivalent to learning the conditional distribution of labels given the input features} in PD learning. 

\subsection{Loss function design with incorporated prior knowledge}
It is crucial to note that the distribution fitting error $d_\Phi(\hat{q}_{\mathcal{Y}|x}, f_\theta(x))$ measures the discrepancy between the target conditional distribution $\hat{q}_{\mathcal{Y}|x}$ and the transformed prediction $(f_\theta(x))_{\Phi^*}^* = \nabla \Phi^*(f_\theta(x))$, rather than directly comparing $\hat{q}_{\mathcal{Y}|x}$ with the raw model output $f_\theta(x)$. This transformation ensures that the model's prediction is properly mapped to the probability simplex or another appropriate domain, depending on the choice of $\Phi$. The purpose of this design is to incorporate prior knowledge $C$ about the target distribution $\hat{q}_{\mathcal{Y}|x}$ into the learning process by carefully choosing the generating function $\Phi$. 
Specifically, given that $q_{\mathcal{Y}|x} \in C$, where $C \subseteq \Delta$ is a convex set encoding constraints on $q_{\mathcal{Y}|x}$, we assume $C$ is non-empty, closed, and convex to ensure well-posedness. We then define the extended convex function:
\[
    \Phi(q_{\mathcal{Y}|x}) = \Omega(q_{\mathcal{Y}|x}) + I_C(q_{\mathcal{Y}|x}),
\]
    where $\Omega$ is a differentiable, strictly convex function. This ensures $\mathrm{dom}(\Phi) \subseteq C$. 
According to Lemma~\ref{lem:sum_conjugate}, we have
\begin{equation}
\Phi^*(\mu) = \inf_{\mu' \in \mathbb{R}^d} \sigma_C(\mu') + \Omega^*(\mu - \mu'),
\end{equation}
where $I_C^*(\mu) = \sigma_C(\mu) = \max_{q_{\mathcal{Y}|x} \in C} \langle \mu, q_{\mathcal{Y}|x} \rangle$. When $\Phi^*(f_\theta(x))$ is finite and differentiable, the Legendre-Fenchel duality implies $f_\theta(x)_{\Phi^*}^* \in \mathrm{dom}(\Phi) \subseteq C$. Thus, the model output $f_\theta(x)_{\Phi^*}^*$ is naturally constrained within $C$, without requiring explicit projection or regularization. 
To illustrate, consider $\Omega(\mu) = \sum_{i=1}^d (\mu_i \log \mu_i - \mu_i)$ (negative Shannon entropy). Its conjugate is $\Omega^*(\nu) = \sum_{i=1}^d e^{\nu_i}$, yielding an unnormalized prediction $[e^{\nu_i}]_{i=1}^d$. 
However, with the probability simplex constraint $C = \{ \mu \in \mathbb{R}^d_+ \mid \sum_i \mu_i = 1 \}$, the conjugate becomes $\Phi^*(\nu) =(\Omega+I_C)^*(\nu)= \log \sum_{i=1}^d e^{\nu_i}$, and the model's prediction becomes $ \nu^*_{\Phi^*} = [ {e^{\nu_i}}/{\sum_{j=1}^d e^{\nu_j}} ]_{i=1}^d$, i.e., the standard Softmax. This shows that Softmax is not merely a heuristic but a natural consequence of using negative entropy and the probability simplex constraint. 
Although it is feasible to directly use $f_\theta(x)$ to fit $q_{\mathcal{Y}|x}$ without this approach, doing so fails to fully utilize the available prior information. This results in a larger search space, reduced training efficiency, and, more critically, provides no guarantee that $f_\theta(x)$ will satisfy the constraint set $C$.  
Therefore, by employing Legendre-Fenchel duality, we introduce a method for designing loss functions that effectively incorporates prior knowledge.  This principle enables the PD learning framework to flexibly adapt to different learning scenarios by encoding domain-specific assumptions directly into the geometry of the loss function, thereby improving both theoretical interpretability and practical performance. For notational convenience, unless otherwise specified, $\Phi$ in the subsequent sections of this paper refers to the loss function constructed via Legendre-Fenchel duality that explicitly incorporates the prior knowledge $C$.

\section{Optimization of PD learning}
\label{sec:optimization}
This section presents an analysis of deep learning through the lens of PD estimation, using the Fenchel-Young loss as the training objective and the learning error as the primary optimization target. 

\subsection{Extended convexity and extended smoothness}
\label{subsec:definition}
Before introducing the proposed method, we first define the concepts of extended convexity and extended smoothness, along with a discussion of their properties. 

We begin with the formal definition of a norm power function.
\begin{definition}[$L_2$-norm power function]
A function $\Psi: \mathbb{R}^m \to \mathbb{R}$ is said to be a norm power function of order $r_{\Psi}$ with scale $a_{\Psi}$, if it can be expressed as:
\begin{equation}
    \Psi(\mu) = \|\mu a_{\Psi}\|_2^{r_\Psi}/r_{\Psi} ,
\end{equation}
where $\|\cdot\|_2$ represents $L_2$ norm, and $r_\Psi > 1$, $a_\Psi>0$. 
\end{definition}

The following lemma summarizes key properties of norm power functions and their normalized counterparts.
\begin{lemma}
\label{lem:prop_thomo_fun}
If $\Psi(\mu)$ is a $L_2$-norm power function, then 
\begin{enumerate}
\item \label{lem:prop_thomo_fun:1} $\Psi^*(\nu)$ is also a norm power function, where
\begin{equation}
  \begin{aligned}
      &1/r_{\Psi} + 1/r_{\Psi^*} = 1,\\
      &a_{\Psi}a_{\Psi^*}=1.
  \end{aligned}
\end{equation}

    \item \label{lem:prop_thomo_fun:2} The relationship between $\Psi$ and its conjugate is given by:
    \begin{equation}
        r_\Psi \Psi(\mu) = r_{\Psi^*} \Psi^*(\mu^*_\Psi) = \mu^\top \mu_\Psi^*,
    \end{equation}
    where $\mu^*_\Psi$ is the dual variable associated with $\mu$. 

\end{enumerate}
\end{lemma}

The proof of this lemma can be found in Appendix~\ref{appendix:proof_prop_thomo_fun}.

\begin{definition}
\label{def:g_convex_smooth}
Define $S_\Phi(\mu,\nu) = \Phi(\mu) - \Phi(\nu) - \langle \nabla \Phi(\nu), \mu - \nu \rangle$.  We define the following extended convexity and smoothness:

\begin{itemize}
    \item \textit{$\mathcal{H}(\psi)$-convex function}. A real-valued function $\Phi$ is said to be $\mathcal{H}(\psi)$-convex if it satisfies:
    \begin{equation}
        S_\Phi(\mu, \nu) \ge \psi(\mu - \nu), \quad \forall \mu, \nu \in \mathrm{dom}(\Phi)^m,
    \end{equation}
    where $\psi$ is a norm power function.

    \item \textit{$\mathcal{H}(\Psi)$-smooth function}. A real-valued function $\Phi$ is said to be $\mathcal{H}(\Psi)$-smooth if it satisfies:
    \begin{equation}
        S_\Phi(\mu, \nu) \le \Psi(\mu - \nu), \quad \forall \mu, \nu \in \mathrm{dom}(\Phi).
    \end{equation}
\end{itemize}
\end{definition}

A function $\Phi$ that is Lipschitz smooth corresponds to an $\mathcal{H}(\Psi)$-smooth function with $r_\Psi=2$. 
Strongly convex functions and higher-order strongly convex functions~\cite{Lin2003SomeEP} are specific instances of $\mathcal{H}(\psi)$-convex functions. In particular, for strongly convex functions $r_\psi=2$. 
This makes them particularly well-suited for modeling practical deep learning scenarios, where loss landscapes are often neither convex nor smooth in the traditional sense.
\textit{When a function is both $\mathcal{H}(\psi)$-convex and $\mathcal{H}(\Psi)$-smooth, the Fenchel–Young loss induced by this function is effectively equivalent to a power of the norm function.} This equivalence facilitates the analysis of optimization dynamics in models relying on such losses.

The following lemma further provides several criteria for determining the extended convexity and smoothness of a function.

\begin{lemma}
\label{lem:convex_smooth}
    \par 1. Suppose $\Phi$ is a twice-differentiable function whose Hessian matrix is positive definite over its domain. Let $\lambda_{\min}$ and $\lambda_{\max}$ denote the smallest and largest eigenvalues of the Hessian matrix $\nabla^2\Phi(\xi)$ for all $\xi$ in the domain of $\Phi$.  Then $\Phi$ is both $\mathcal{H}(\lambda_{\max}\|\cdot\|_2^2/2)$-smooth and $\mathcal{H}(\lambda_{\min}\|\cdot\|_2^2/2)$-convex.
    \par 2. Let $s$ be a fixed constant vector. If $\Phi$ is an extended convex function, then $d_\Phi(\mu, s)$ has the same extended convexity and smoothness properties as $\Phi(\mu)$, and $d_\Phi(s, \mu)$ has the same extended convexity and smoothness properties as $\Phi^*(\mu)$.
    \par 3. In practical scenarios, for the loss function $d_\Phi(y, f_\theta(x))$ associated with any neural network model $f_\theta(x)$, there exist $L_2$-norm power functions $\psi$, $\Psi$, and $\xi$ such that the loss function is $\mathcal{H}(\psi)$-convex and $\mathcal{H}(\Psi)$-smooth with respect to the model output $f_\theta(x)$, and $\mathcal{H}(\xi)$-smooth with respect to the parameter $\theta$.
\end{lemma}
The proof of this lemma is provided in Appendix~\ref{appendix:proof_convex_smooth}. 
The Conlution 2 of lemma~\ref{lem:convex_smooth} demonstrates that the extended smoothness and convexity of the Fenchel-Young loss $d_\Phi(\mu, s)$ can be analyzed through its generating function $\Phi$. Therefore, the MSE loss is both $\mathcal{H}(\|\cdot\|_2^2)$-convex and $\mathcal{H}(\|\cdot\|_2^2)$-smooth. Based on Conclusion 1 of Lemma~\ref{lem:convex_smooth}, the Softmax Cross-Entropy loss and its equivalent KL divergence $D_{\mathrm{KL}}(s, \mathrm{Softmax}(f_\theta(\mathcal{X})))$ are both $\mathcal{H}(\psi)$-convex and $\mathcal{H}(\Psi)$-smooth with respect to $\mathrm{Softmax}(f_\theta(\mathcal{X}))$, where $r_{\psi}=r_{\Psi}=2$. The proposed notions of $\mathcal{H}(\psi)$-convexity and $\mathcal{H}(\Psi)$-smoothness are capable of describing optimization objectives in deep learning in a more unified and realistic manner, thereby offering a convenient tool for analyzing non-convex optimization problems.

\begin{proposition}
\label{prop:h_bound}
Let $G_* = \min_{\mu} G(\mu)$, $\mathcal{G}_* = \{\mu|G(\mu)=G_*\}$, and $\mu_* \in \mathcal{G}_*$. 
The following results hold:
\par 1. If $G(\mu)$ is $\mathcal{H}(\psi)$-smooth, then:
\begin{equation}
    \Psi^*(\mu_G^*)  \leq G(\mu) - G_* \leq \Psi(\mu - \mu_*) .
\end{equation}
\par 2. If $G(\mu)$ is $\mathcal{H}(\psi)$-convex, then:
\begin{equation}
    \psi(\mu - \mu_*)  \leq G(\mu) - G_* \leq \psi^*(\mu_G^*) .
\end{equation}

\par 3. If $G(\mu)$ is both $\mathcal{H}(\psi)$-smooth and $\mathcal{H}(\psi)$-convex, then:
\begin{equation}
    \Psi^*(\mu_G^*)  \leq G(\mu) - G_* \leq \psi^*(\mu_G^*).
\end{equation}

\end{proposition}
The proof of this lemma is provided in Appendix~\ref{appendix:proof_h_bound}.
The above proposition reveals that, in addition to their elegant symmetry, $\mathcal{H}(\psi)$-convexity and $\mathcal{H}(\psi)$-smoothness establish a strong and intuitive connection between the global optimum of a function and its gradient energy. 

\subsection{Convergence of SGD under Extended Smoothness}

In this section, we establish the convergence of SGD under the extended smoothness condition. 
For notational convenience, let $s_k$ denote a mini-batch sampled from the dataset $s$, and define $G(\theta, s_k) = \frac{1}{|s_k|} \sum_{z \in s_k} G(\theta, z)$, where $|s_k|$ is the number of samples in the mini-batch. Consider the optimization problem
\begin{equation}
    \min_{\theta}  G(\theta, s) = \min_{\theta} \mathbb{E}_Z G(\theta, Z),
\end{equation}
where $Z \sim \hat{q}$ and the objective function $G: \mathbb{R}^m \to \bar{\mathbb{R}}$ is $\mathcal{H}(\xi)$-smooth with respect to $\theta$ for $z\in \mathcal{Z}$. Denote $G_* = \min_{\theta} G(\theta,s)$, $\mathcal{G}_* = \{\theta \mid G(\theta,s) = G_*\}$, and let $\theta_* \in \mathcal{G}_*$. 
The update rule of SGD is given by
\begin{equation}
    \theta_{k+1} = \theta_k - \alpha g(\theta_k, s_k),
\end{equation}
where $g(\theta_k, s_k)$ is the stochastic gradient at iteration $k$; $\theta_k$ is the parameter value at iteration $k$; and $s_k$ is the mini-batch sampled from $s$.

The following proposition characterizes the convergence behavior of SGD under $\mathcal{H}(\xi)$-smoothness.
\begin{proposition}
\label{prop:sgd_convergence}
Suppose $G(\theta,z)$ is an $r_\xi$-degree $\mathcal{H}(\xi)$-smooth function satisfying the following conditions:
Then the following statements hold:
    \par 1. The optimal learning rate is achieved when $\alpha = \left(\frac{\|g(\theta_k,s_k)\|_2^2}{r_\xi \xi(g(\theta_k,s_k))}\right)^{{1}/{r_\xi-1}}$. Under this choice, the following inequality holds:
    \begin{equation}
        G(\theta_{k+1},s_k) \leq G(\theta_{k},s_k) - r_{\xi^*}^{-1} \left(\frac{\|g(\theta_k,s_k)\|_2^2}{\bar{\xi}(g(\theta_k,s_k))}\right)^{r_{\xi^*}},
    \end{equation}
    where $r_{\xi^*} = r_\xi / (r_\xi - 1)$.
    \par 2.  If the mini-batch $s_k$ is selected such that
    \begin{equation}
        G(\theta_{k+1},s) - G(\theta_k,s) \leq \beta \big( G(\theta_{k+1},s_k) - G(\theta_k,s_k) \big)
    \end{equation}
    for some constant $\beta > 0$, then to achieve $\mathbb{E} \|\nabla G(\theta_k,s_k)\|_2^{r_{\xi^*}} \leq \varepsilon^{r_{\xi^*}}$, the required number of iterations $T$ satisfies $T = \mathcal{O}\left(\varepsilon^{-r_{\xi^*}}\right)$\label{eq:sgd_complexity}. 
\end{proposition}
The proof of this lemma is provided in Appendix~\ref{appendix:proof_sgd_convergence}.

\subsection{Non-convex optimization mechanism}\label{subsec:logic}

In this subsection, we analyze the process by which the empirical risk is minimized during training. 

\begin{proposition}
   \label{prop:st_H_eigenvalue_bound}
   
   Let Let $L_2^2(\cdot) = \|\cdot\|_2^2$, $g(x,y)=\nabla_\theta d_{\Phi}(\mathbf{1}_y,f_\theta(x))$. If $\lambda_{\min}(A_s)\neq 0$, then the following statements hold:

    \par 1. \textit{Individual Sample Bound:} 
    \begin{equation}
    \begin{aligned}
        \frac{\|g(x,y)\|_2^2}{\lambda_{\max}(A_x)} &\leq \|\mathbf{1}_y-f_\theta(x)_{\Phi^*}^*\|_2^2 \leq \frac{\|g(x,y)\|_2^2}{\lambda_{\min}(A_x)}.
    \end{aligned}
    \end{equation}
    \par 2. \textit{Dataset Bound:} 
    \begin{equation}
    \begin{aligned}
     \mathcal{L}_{L_2^2/2}(\hat{q},f_\theta(\cdot)_{\Phi^*}^*) \leq \mathbb{E}[{\|g(X,Y)\|_2^2}/(2\lambda_{\min}(A_X))],\\
       \mathcal{L}_{\Phi}(\hat{q},f_\theta(\cdot)) 
           \le \mathbb{E}_{XY\sim \hat{q}} \psi^*(g(X,Y))/\lambda_{\min}(A_s)^{r_{\psi^*}/2}.
    \end{aligned}
    \end{equation}
\end{proposition}
The proof can be found in Appendix~\ref{appendix:proof_st_H_eigenvalue_bound}.

The above proposition indicates that $\mathcal{L}_{\Phi}(\hat{q},f_\theta(\cdot))$ can be optimized by reducing the gradient energy and increasing the eigenvalues of the structure matrix. Since $d_\Phi(\mathbf{1}_y,f_\theta(x))$ is $\mathcal{H}(\xi)$-smooth with respect to $\theta$, according to Proposition~\ref{prop:sgd_convergence}, the SGD algorithm can reduce the gradient energy, thereby reducing the $\mathcal{L}_{\Phi}(\hat{q},f_\theta(\cdot))$ and $\mathcal{L}_{L_2^2/2}(\hat{q},f_\theta(\cdot)_{\Phi^*}^*)$.  
Another key factor influencing the empirical risk is the extreme eigenvalues of the structure matrix. In the study~\cite{Qi2025}, the mechanisms by which skip connections, random parameter initialization, and overparameterization control these extreme eigenvalues have been theoretically and experimentally demonstrated. 
In summary, by extending the concepts of strong convexity and Lipschitz smoothness, we have proven that, under the PD learning framework, gradient energy and empirical risk are equivalent. Furthermore, the SGD algorithm reduces gradient energy, while model architectures and other deep learning techniques can control the extreme eigenvalues of the structure matrix. This provides theoretical support for the non-convex optimization mechanisms in PD learning, and more broadly, in deep learning problems.

\section{Generalization Bounds in PD learning}
\label{sec:bounds}
In this subsection, we will further discuss the generalization capability of models in PD learning. 

\subsection{Bounds on risk and generalization error}
We first establish the model-independent upper and lower bounds on the expected and empirical risks for a trained model as follows:
\begin{proposition}
    \label{prop:risk_bound}
    The expected and empirical risks under the PD learning framework are bounded as follows:
    \begin{equation}
    \begin{aligned}
           0 &\leq \min_{\theta} \mathcal{L}_{\Phi}(\hat{q},f_\theta(\cdot))-\mathrm{Ent}_\Phi(\mathbf{1}_Y|X) \leq \mathrm{Ent}_\Phi(\hat{q}_{\mathcal{Y}|X}), \\
        0 &\leq \min_{\theta} \mathcal{L}_{\Phi}(q,f_\theta(\cdot)) -\mathrm{Ent}_\Phi(\mathbf{1}_{\hat{Y}}|\hat{X})\leq \mathrm{Ent}_\Phi(q_{\mathcal{Y}|X}).
    \end{aligned}
   \end{equation}
\end{proposition}
The proof of this proposition is provided in Appendix~\ref{appendix:proof_risk_bound}. 
The bounds on the empirical risk in this proposition provide a theoretical justification for our statement: \textit{Data determines the upper and lower bounds of fitting and learning, while the model determines how closely these bounds are approached.} The lower bound is the generalized conditional entropy ($\mathrm{Ent}_\Phi(\mathbf{1}_Y|X)$)~\ref{def:generalized_entropy}, while the upper bound is the sum of this entropy and the generalized mutual information $\mathrm{Ent}_\Phi(q_{\mathcal{Y}|X})$. The generalized mutual information $\mathrm{Ent}_\Phi(q_{\mathcal{Y}|X})$ is essentially a non-negative measure of the dependence between $X$ and $Y$. This becomes particularly intuitive when $\Phi$ is chosen as the negative Shannon entropy. By setting $\Phi$ to the negative Shannon entropy, we can connect the above bounds directly to classical Shannon information theory as follows.
\begin{corollary}
\label{cor:kl_mutual_info_bound}
For $\Phi(q) = \sum_i q_i \log q_i$, we have:
\begin{equation}
    \label{eq:info_bound}
    H(\hat{Y}|\hat{X}) \leq \min_{\theta} \mathcal{L}_{\Phi}(\hat{q},f_\theta(\cdot)) \leq I(\hat{X}; \hat{Y}) + H(\hat{Y}|\hat{X}),
\end{equation}
where $I(\hat{X};\hat{Y})$ is the Shannon mutual information, $H(\hat{Y}|\hat{X})$ is the Shannon conditional entropy, and $(\hat{X}, \hat{Y}) \sim \hat{q}$.
\end{corollary}
This corollary leads to an interesting insight: the smaller the mutual information between features and labels, the tighter the upper bound on the empirical risk of the trained model. In other words, the mutual information between features and labels under the empirical distribution determines the required model capacity, and smaller mutual information implies that simpler models achieve lower empirical risk, meaning the data is easier to fit.

We now analyze the generalization error within the PD learning framework. To begin, we first define the model-induced information loss in PD learning.
\begin{definition}
Let $f_\theta(\mathcal{X}) = \{f_\theta(x) \mid \forall x \in \mathcal{X}\}$ be the image of the input space $\mathcal{X}$ under $f_\theta$. We define the information loss induced by the model $f_*$ as $\zeta = |\mathcal{X}| - |f_\theta(\mathcal{X})|$.
\end{definition}
When $|\mathcal{X}|$ is finite, $f_\theta$ acts as a function mapping from $\mathcal{X}$ to its output space, and thus $\zeta \geq 0$. The quantity $\zeta$ essentially measures the reduction in the number of distinct support points in the output space after mapping through $f_\theta$. In other words, it quantifies the degree of information collapse, how many distinct inputs are mapped to the same output value. We now present a probabilistic bound that characterizes the impact of model-induced information loss, sample size, and maximum loss on the generalization performance in the PD learning framework.

\begin{proposition}
    \label{prop:prob_gen_bound}
Let $f_*$ denote the trained model, and let $\gamma = \max_{(x,y) \in \mathcal{X} \times \mathcal{Y}} d_\Phi(\mathbf{1}_y, f_*(x))$. Then, we have:  $\forall \varepsilon \ge \gamma \sqrt{\frac{5(|\mathcal{X}|-\zeta)|\mathcal{Y}|}{n}}$
\begin{equation}
\begin{aligned}
            &\Pr\left( \left| \mathcal{L}_{\Phi}(q,f_*(\cdot)) - \mathcal{L}_{\Phi}(\hat{q},f_*(\cdot)) \right| \ge \varepsilon \right) 
        \le 3\exp\left(-\frac{4n\varepsilon^2}{25\gamma^2}\right).
\end{aligned}
    \end{equation}
\end{proposition}
The proof of this proposition is provided in Appendix~\ref{appendix:proof_prob_gen_bound}.
Based on this proposition, we identify three controllable factors that influence the generalization error:
\par 1. A larger sample size $n$ makes the generalization error easier to control. This observation is consistent with the implications of the central limit theorem and with generalization error bounds derived from Hoeffding's inequality.
\par 2. The quantity $\gamma$ is a key factor in controlling the upper bounds on the expected risk and generalization error. In particular, when $\Phi(\cdot)$ is the negative Shannon entropy, we have $\gamma = -\log p_{\min}$, where $p_{\min}$ is the smallest predicted probability for any label. In other words, when the model uses KL divergence or Cross-Entropy loss, a smoother output distribution (e.g., from the softmax function) with a larger minimum predicted probability leads to a tighter upper bound on the generalization error.
\par 3. A larger value of $\zeta$ implies more information loss and greater compression, which is detrimental to controlling generalization error. Therefore, we arrive at a concise and powerful conclusion: \textit{increasing the information loss induced by the model helps control the generalization error.}

\subsection{Regularization and generalization}
We now use the conclusions from the above proposition to explain the impact of regularization terms on model generalization. First, we establish a result on the equivalence between parameter norm and the smoothness of model outputs.
\begin{proposition}
\label{thm:regularization_generalization}
Given that for all $ x \in \mathcal{X} $, $\Phi^*(f_\mathbf{0}(x))=\min_\theta \Phi^*(f_{\theta}(x))$, it follows that there exist positive constants $a$, $b$, and $\epsilon$ such that: When $\theta \in \Theta'_\epsilon$, the norm $\|\theta\|_2$ controls the behavior of the function $\Phi^*(f_{\theta}(x))$ as follows:
\begin{equation}
        a\|\theta\|_2^2 \leq \Phi^*(f_{\theta}(x)) - \Phi^*(\mathbf{0}) \leq b\|\theta\|_2^2.
\end{equation}
\end{proposition}
The proof of this result is provided in Appendix~\ref{appendix:proof_regularization_generalization}.

The assumption $f_{\mathbf{0}}(x)=\mathbf{0}$ and $\Phi^*(\mathbf{0}) = \min_{\theta} \Phi^*(f_\theta(x))$  is well-motivated by standard practices in DNNs. First, when all model parameters are set to zero (i.e., $\theta = \mathbf{0}$), it is common for most DNNs to satisfy $f_{\theta}(x) = \mathbf{0}$. This behavior arises naturally in widely used architectures such as fully connected layers with zero-setting weights and biases, convolutional layers with zero-setting kernels, and transformer modules. In each of these cases, setting all parameters to zero results in outputs that are zero vectors, making the assumption both reasonable and practically relevant. Therefore, the assumption in Proposition~\ref{thm:regularization_generalization} is both realistic and consistent with standard practices in neural network design.  Below, we analyze the impact of label smoothing and parameter regularization on the maximum value of the loss function.
\begin{proposition}
    \label{prop:generalize_explain}
If $f_{\mathbf{0}}(x) = \mathbf{0}$ for all $x \in \mathcal{X}$ and $\mathbf{0}_{\Phi^*} = u$, where $u$ is the uniform distribution on $\mathcal{Y}$, then reducing $\|\theta\|_2^2$ for all $x \in \mathcal{X}$ is equivalent to reducing $\gamma$.
\end{proposition}
The proof of this lemma can be found in Appendix~\ref{appendix:proof_generalize_explain}. 
According to Proposition~\ref{prop:prob_gen_bound}, reducing $\gamma$ helps in controlling the generalization error. As shown in Proposition~\ref{prop:generalize_explain}, the use of $L_2$-norm regularization can effectively reduce $\gamma$. From this perspective, in deep learning and more broadly in machine learning, \textit{the mechanism by which label smoothing and parameter regularization help control generalization error is achieved through the control of $\gamma$}.

\section{Empirical validation}
\label{sec:experiments}

In this section, we aim to validate the correctness of the core theoretical conclusions of this paper through experiments.  
\begin{figure}[htbp]
\begin{center}
\centerline{\includegraphics[width=\columnwidth]{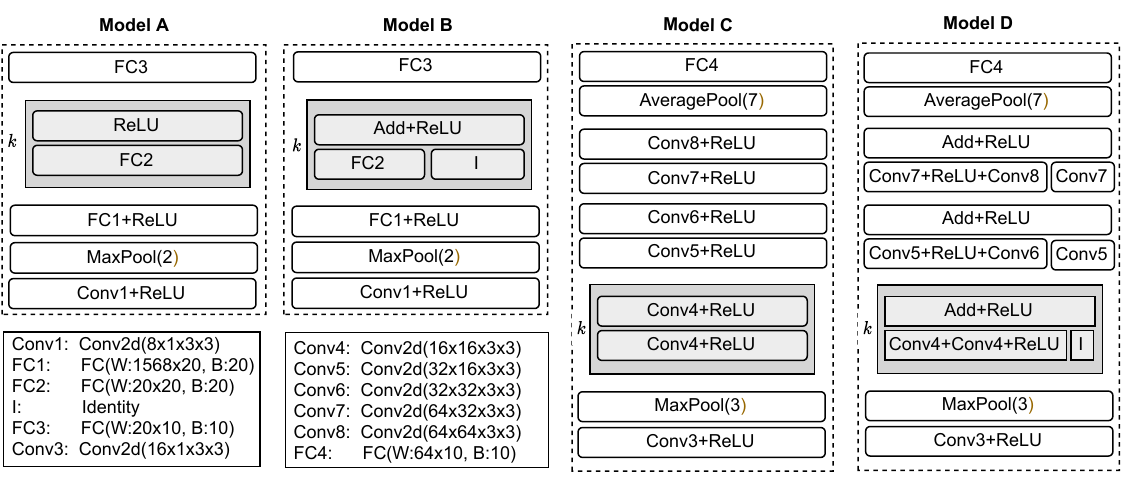}}
\caption{Model architectures and configuration parameters. The gray blocks in the diagram indicate components whose number of iterations can be adjusted via the parameter $k$. Model B and Model D are variants of Model A and Model C, respectively, incorporating skip connections. The symbol $I$ denotes an identity function transformation.}
\label{fig:model_increase}
\end{center}
\vskip -0.2in
\end{figure}
\subsection{Experimental design}

\textit{Dataset:} We use the \texttt{MNIST}~\cite{LeCun1998GradientbasedLA}, \texttt{FashionMNIST}~\cite{DBLP:journals/corr/abs-1708-07747}, \texttt{CIFAR-10}, and \texttt{CIFAR-100}~\cite{krizhevsky2009learning} datasets for evaluation. 

\textit{Models:} We employ two sets of models tailored to the specific theoretical properties being verified: one consisting of our custom-designed models and the other comprising widely studied, classical model architectures. 
Our custom-designed model architectures and configuration parameters are illustrated in Fig.~\ref{fig:model_increase}. Models B and D extend Models A and C by incorporating skip connections, with $k$ denoting the number of repeated blocks, allowing for adjustable model depth. Specifically, $k$ represents the number of times the corresponding block in Fig.~\ref{fig:model_increase} is iterated; for instance, $k=2$ indicates that two identical blocks are cascaded sequentially. As $k$ increases, the number of cascaded blocks grows, resulting in a higher total number of trainable parameters. 
The classical model architectures include: LeNet~\cite{LeCun1998GradientbasedLA}, GoogLeNet~\cite{DBLP:conf/cvpr/SzegedyLJSRAEVR15}, ResNet18~\cite{He2015DeepRL}, PreActResNet18~\cite{DBLP:conf/eccv/HeZRS16}, DenseNet121~\cite{DBLP:conf/cvpr/HuangLMW17}, MobileNet~\cite{DBLP:journals/corr/HowardZCKWWAA17}, ShuffleNetV2~\cite{DBLP:conf/eccv/MaZZS18}, SENet18~\cite{DBLP:journals/pami/HuSASW20}, Vision Transformers (ViT)~\cite{dosovitskiy2021an}, DynamicTanh (DyT)~\cite{DBLP:conf/cvpr/0002CHL025}, and Class-Attention in Image Transformers (CaiT)~\cite{DBLP:conf/iccv/TouvronCSSJ21}. The architectural details of these models can be found in the work~\cite{yoshioka2024visiontransformers}.

\textit{Environment and Hyperparameters:} The experiments were conducted in the following computing environment: Python 3.7, PyTorch 2.2.2, and a GeForce RTX 2080 Ti GPU.  
The training configuration is as follows: the loss function is Softmax Cross-Entropy, the batch size is 64, and optimization is performed using SGD with a learning rate of 0.01 and momentum of 0.9. 
For our custom-designed models trained on the relatively simple \texttt{MNIST} and \texttt{Fashion-MNIST} datasets, we perform only one epoch of training, while controlling model capacity by varying the hyperparameter $k$. For the more complex classical architectures, adopted from the literature and evaluated on \texttt{CIFAR-10} and \texttt{CIFAR-100}, we set the training epoch count to 40.  
\subsection{Experimental results on the optimization mechanism}
In the experiments, we employ the Softmax Cross-Entropy loss to guide the model's predictions toward the labels in the training samples. We have established that the Cross-Entropy loss and its equivalent KL divergence are both $\mathcal{H}(\psi)$-convex and $\mathcal{H}(\Psi)$-smooth with respect to $p =f_\theta(x)_{\Phi^*}^*= \mathrm{Softmax}(f_\theta(x))$. This implies that $D_{\mathrm{KL}}(\mathbf{1}_y, \mathrm{Softmax}(f_\theta(x)))$ behaves similarly to $\|\mathbf{1}_y - \mathrm{Softmax}(f_\theta(x))\|_2^2$. To mitigate the numerical instability and large fluctuations in distance caused by the $\log$ function in the KL divergence when probabilities approach zero, we use the equivalent functional form $\mathrm{D}_{L^2_2/2}(\mathbf{1}_y,f_\theta(x)_{\Phi^*}^*) = \frac{1}{2} \|\mathbf{1}_y - \mathrm{Softmax}(f_\theta(x))\|_2^2$ as a surrogate to represent the model's risk. Specifically, we focus on empirically validating the theoretical bound stated in Proposition~\ref{prop:st_H_eigenvalue_bound}.

\begin{figure}[htbp]
	\centering
	\begin{minipage}{1.0\linewidth}
		\centering
            \centerline{\includegraphics[width=\columnwidth]{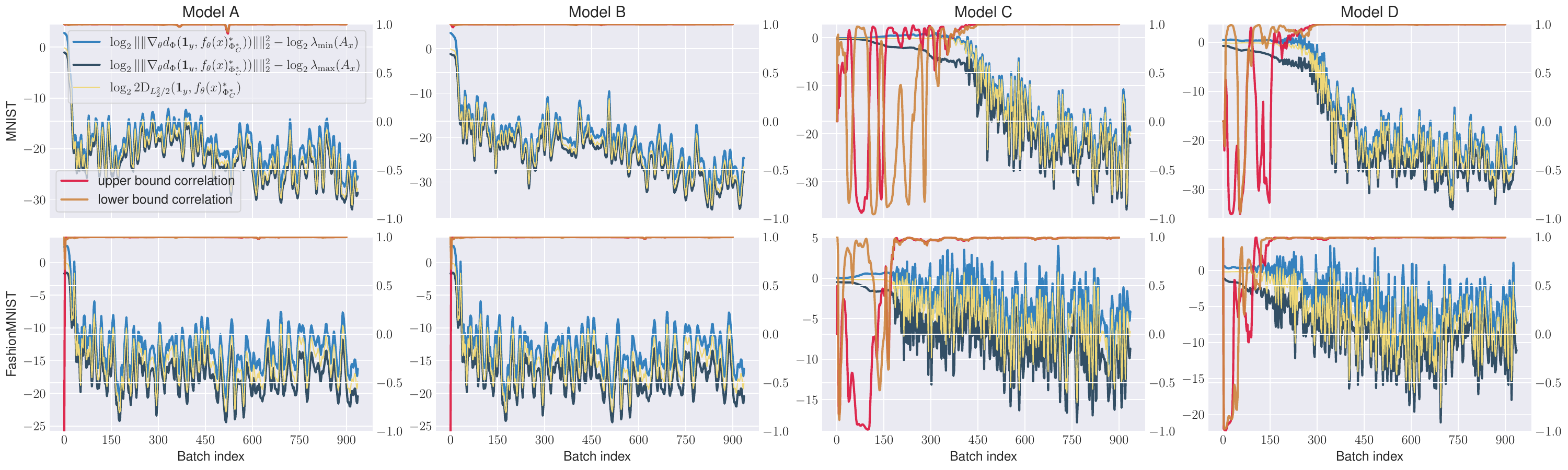}}
            \caption{Training dynamics of the risk along with its upper and lower bounds, and the local Pearson correlation coefficients between them. The red and orange curves represent the local Pearson correlation coefficients, plotted against the right $y$-axis with range $[-1, 1]$. The risk and its bounds are shown on the left $y$-axis.}
            \label{fig:convergence_bound}
            \vspace{5mm} 
	\end{minipage}
	\begin{minipage}{1.0\linewidth}
		\centering
        \centerline{\includegraphics[width=1.0\linewidth]{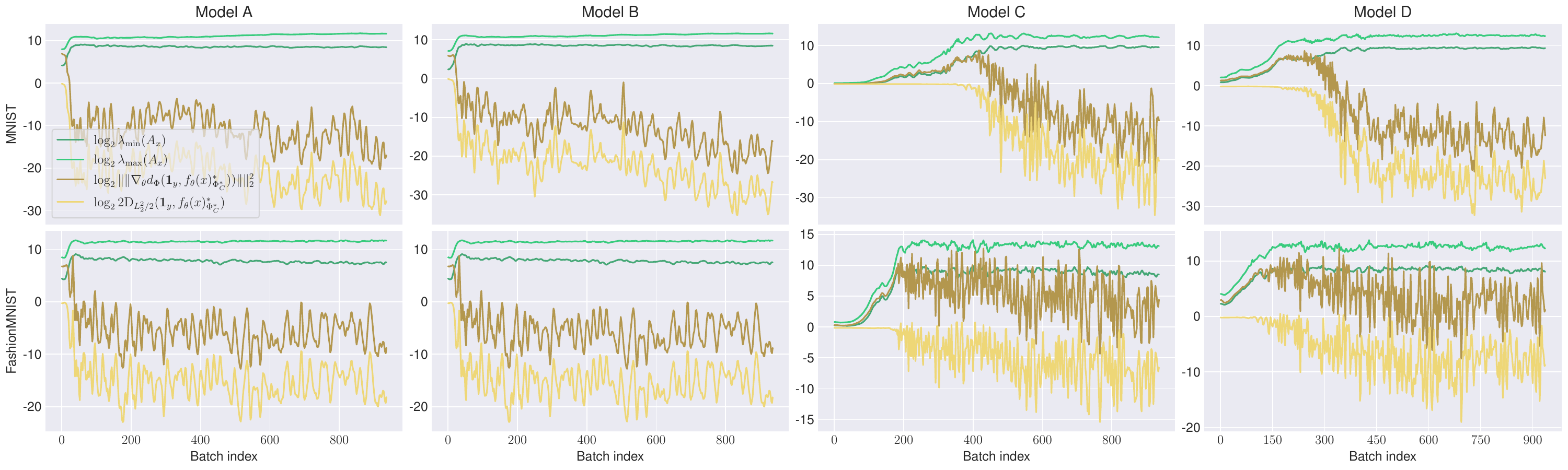}}
        \caption{Training dynamics of the gradient energy and extreme eigenvalues.}
        \label{fig:convergence_indicator}
	\end{minipage}
\vskip -0.1in
\end{figure}
\begin{figure}[htbp]
\centering
    \centerline{\includegraphics[width=1.0\linewidth]{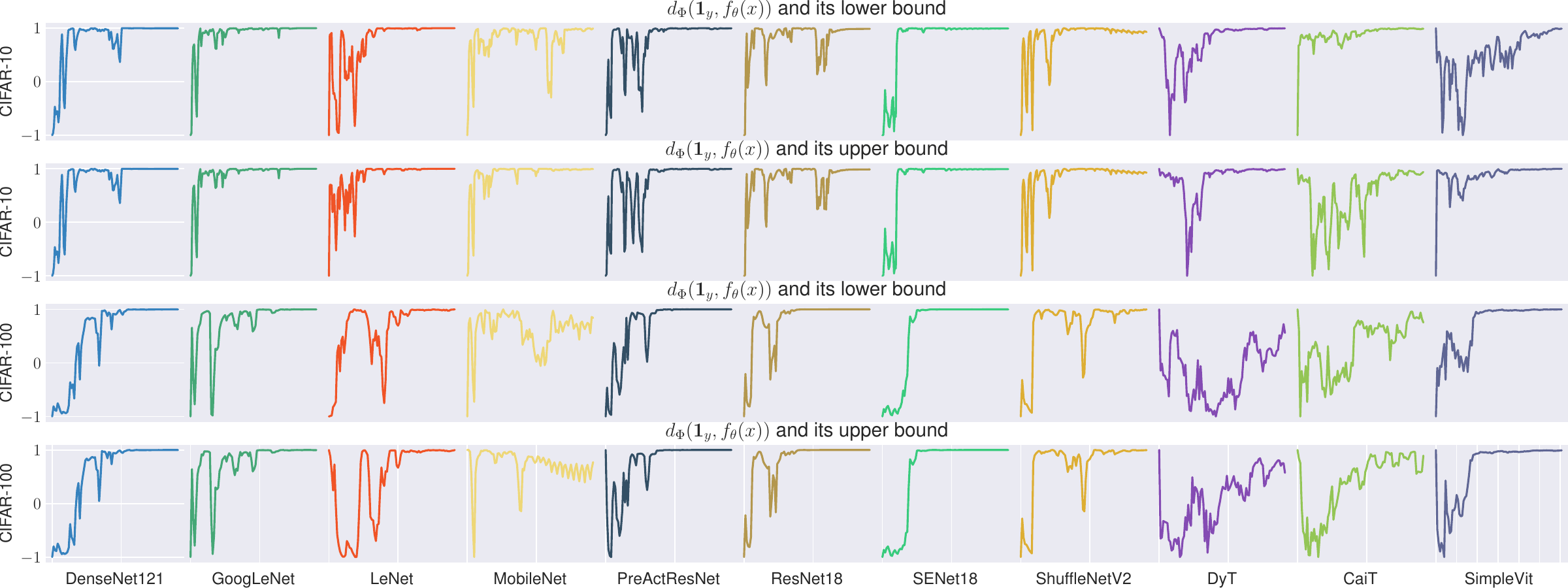}}
    \caption{Training dynamics of the local Pearson correlation coefficient on CIFAR-10 and CIFAR-100. In each subplot, the horizontal axis represents the epoch index, with values ranging from 0 to 40, while the vertical axis denotes the local Pearson correlation coefficient.}
    \label{fig:diff_net_cifar}
\end{figure}

\begin{figure}[htbp]
	\centering
    \centerline{\includegraphics[width=0.9\linewidth]{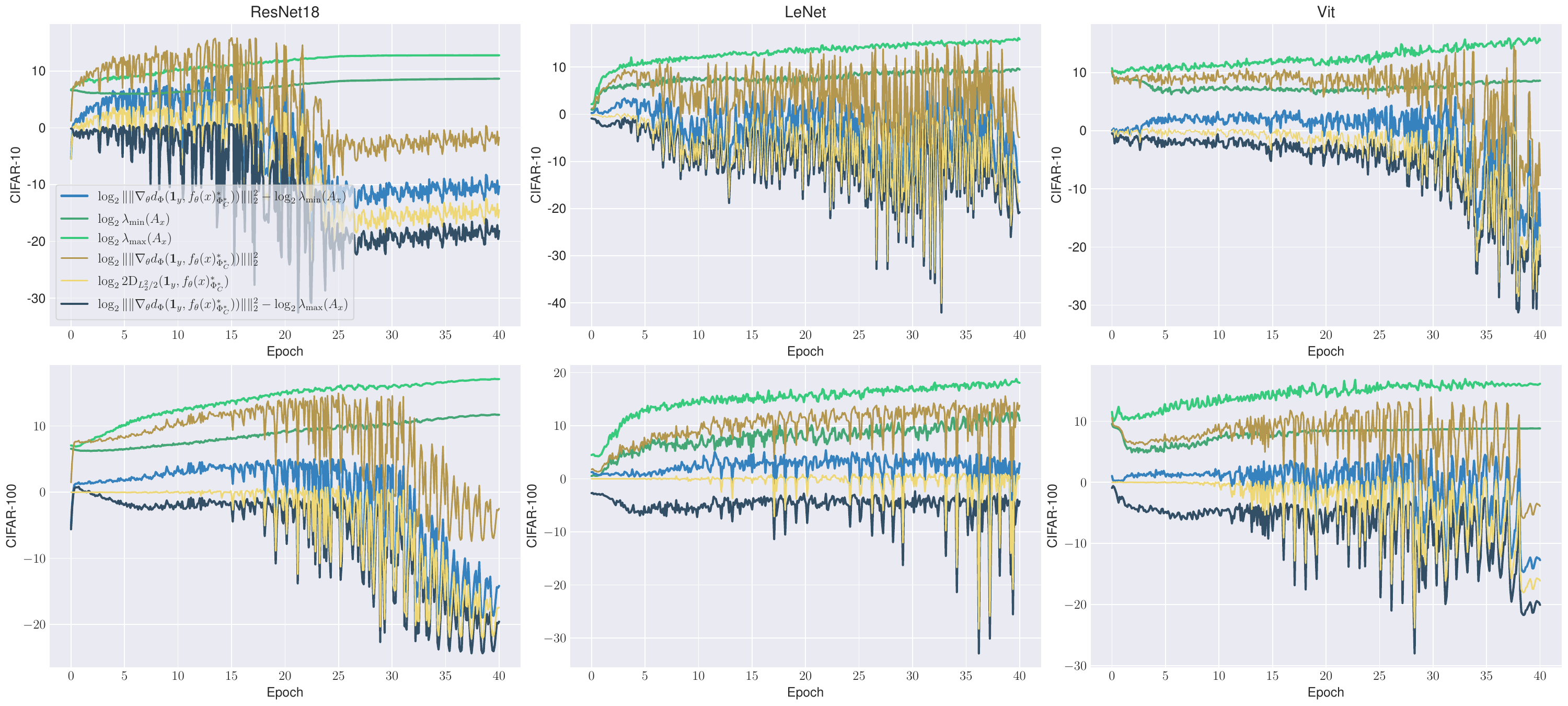}}
    \caption{Training dynamics on the CIFAR-10 and CIFAR-100 datasets.}
    \label{fig:classic_net_info}
\end{figure}

For the custom-designed models evaluated on the MNIST and Fashion-MNIST datasets, we set $k = 1$ for all models in Fig.~\ref{fig:model_increase} and train them under this configuration.
The risks of the models, along with their respective bounds, are depicted in Fig.~\ref{fig:convergence_bound}. To quantify the changes in correlations among these metrics as training progresses, we introduce the local Pearson correlation coefficient as a metric. Specifically, we apply a sliding window approach to compute the Pearson correlation coefficients between the given variables within each window. We use a sliding window of length 20 to compute the Pearson correlation coefficients between the risk and its upper and lower bounds. During training, the local Pearson correlation coefficients are illustrated in Fig.~\ref{fig:convergence_bound}. We further analyze the changes in extreme eigenvalues of the structure matrix and gradient energy during the training process, as illustrated in Fig.~\ref{fig:convergence_indicator}. 
Beyond MNIST and Fashion-MNIST, on the CIFAR-10 and CIFAR-100 datasets, the local Pearson correlation coefficient curves of classical model architectures are shown in Fig.~\ref{fig:diff_net_cifar}.  
To avoid an overly lengthy presentation of training dynamics across all models, we select LeNet, ResNet-18, and ViT as representative examples of standard convolutional networks, architectures with skip connections, and Transformer-based models, respectively. Fig~\ref{fig:classic_net_info} illustrates the training dynamics of key metrics for these models, namely, risk, its upper and lower bounds, extreme eigenvalues of the structure matrix, and gradient energy. 
Based on the experimental results, we draw the following conclusions:
    \par 1. As shown in Figs.~\ref{fig:convergence_bound},~\ref{fig:classic_net_info}, during training across diverse datasets and model architectures, the quantities $\log_2 \frac{\|\nabla_\theta d_{\Phi}(\mathbf{1}_y, f_\theta(x)_{\Phi^*}^*))\|_2^2}{\lambda_{\min}(A_x)}$ and $\log_2 \frac{\|\nabla_\theta d_{\Phi}(\mathbf{1}_y, f_\theta(x)_{\Phi^*}^*))\|_2^2}{\lambda_{\max}(A_x)}$ serve as upper and lower bounds on the risk $\log_2 2\mathrm{D}_{L^2_2/2}(\mathbf{1}_y, f_\theta(x)_{\Phi^*}^*)$,  respectively. This empirically validates the inequality stated in Proposition~\ref{prop:st_H_eigenvalue_bound}. 
    \par 2. As training progresses, the evolution of the risk increasingly aligns with the trends of its theoretical upper and lower bounds. Simultaneously, the gap between these bounds gradually narrows and approaches zero. Moreover, the local Pearson correlation coefficients between the risk and its upper and lower bounds converge to 1 across all models and datasets as training proceeds. This indicates that the risk is asymptotically governed by its bounds, and thus, effective optimization of the risk is achieved through the control of these bounds.
    \par 3. Figures~\ref{fig:convergence_indicator} and \ref{fig:classic_net_info} reveal the detailed underlying mechanism of this bound control. As training progresses, the gradient energy consistently decreases, a behavior theoretically guaranteed by the SGD algorithm~\ref{prop:sgd_convergence}. Meanwhile, the extreme eigenvalues of the structure matrix generally increase compared to their initial values at random initialization. Once these extreme eigenvalues stabilize, the risk begins to decrease monotonically with the gradient energy. In other words, the stabilization (or convergence) of the extreme eigenvalues of the structure matrix is a prerequisite for effective reduction of the risk through gradient energy minimization.  

\section{Conclusion}
\label{sec:conclusion}

In this paper, we introduce the PD learning framework to elucidate the optimization and generalization mechanisms in deep learning. Within this framework, based on the objective of learning conditional probability distributions, we define fundamental and natural conditions that the loss function, model, and prior knowledge must satisfy.
Building upon these conditions, we first prove theoretically that in PD learning, the loss function has a unique structure and is equivalent to the Fenchel-Young loss. This establishes the general applicability of analytical results based on the Fenchel-Young loss, both in theory and practice.
Second, by extending the concepts of strong convexity and Lipschitz smoothness, we provide theoretical justification for solving the non-convex optimization problem in PD learning using SGD, and we present experimental results that support this theoretical analysis.
Additionally, we establish model-independent upper and lower bounds on the expected risk, empirical risk, and generalization error. In contrast to traditional generalization bounds, our analysis reveals that increasing the information loss induced by the model helps reduce generalization error, and that the mutual information between features and labels determines the required number of model parameters.
We believe that, compared to classical theoretical frameworks, the PD learning framework offers enhanced capability in characterizing the non-convex optimization and generalization mechanisms in deep learning, and holds strong potential for analyzing deep learning problems.

\section*{Acknowledgments}
This work was supported in part by the National Natural Science Foundation of China under Grant Nos. 72171172.


\bibliographystyle{IEEEtran}  
\small
\bibliography{IEEEabrv,reflib}

\begin{thebibliography}{10}
\providecommand{\url}[1]{#1}
\csname url@samestyle\endcsname
\providecommand{\newblock}{\relax}
\providecommand{\bibinfo}[2]{#2}
\providecommand{\BIBentrySTDinterwordspacing}{\spaceskip=0pt\relax}
\providecommand{\BIBentryALTinterwordstretchfactor}{4}
\providecommand{\BIBentryALTinterwordspacing}{\spaceskip=\fontdimen2\font plus
\BIBentryALTinterwordstretchfactor\fontdimen3\font minus \fontdimen4\font\relax}
\providecommand{\BIBforeignlanguage}[2]{{%
\expandafter\ifx\csname l@#1\endcsname\relax
\typeout{** WARNING: IEEEtran.bst: No hyphenation pattern has been}%
\typeout{** loaded for the language `#1'. Using the pattern for}%
\typeout{** the default language instead.}%
\else
\language=\csname l@#1\endcsname
\fi
#2}}
\providecommand{\BIBdecl}{\relax}
\BIBdecl

\bibitem{Ghadimi2013StochasticFA}
S.~Ghadimi and G.~Lan, ``Stochastic first- and zeroth-order methods for nonconvex stochastic programming,'' \emph{{SIAM} J. Optim.}, vol.~23, no.~4, pp. 2341--2368, 2013.

\bibitem{He2015DeepRL}
K.~He, X.~Zhang, S.~Ren, and J.~Sun, ``Deep residual learning for image recognition,'' in \emph{Proc. IEEE Conf. Comput. Vis. Pattern Recognit.}\hskip 1em plus 0.5em minus 0.4em\relax Las Vegas, NV, USA: {IEEE} Computer Society, 2016, pp. 770--778.

\bibitem{Zhang2016UnderstandingDL}
C.~Zhang, S.~Bengio, M.~Hardt, B.~Recht, and O.~Vinyals, ``Understanding deep learning (still) requires rethinking generalization,'' \emph{Commun. {ACM}}, vol.~64, no.~3, pp. 107--115, 2021.

\bibitem{Zhang2021UnderstandingDL}
C.~Zhang, S.~Bengio, M.~Hardt \emph{et~al.}, ``Understanding deep learning (still) requires rethinking generalization,'' \emph{Commun. ACM}, vol.~64, no.~3, pp. 107--115, 2021.

\bibitem{Lee2016GradientDO}
I.~Panageas and G.~Piliouras, ``Gradient descent only converges to minimizers: Non-isolated critical points and invariant regions,'' 2016, \textit{arXiv:1605.00405}.

\bibitem{Jentzen2018StrongEA}
A.~Jentzen, B.~Kuckuck, A.~Neufeld, and P.~von Wurstemberger, ``Strong error analysis for stochastic gradient descent optimization algorithms,'' \emph{Ima Journal of Numerical Analysis}, vol.~41, no.~1, pp. 455--492, 2018.

\bibitem{murty1985some}
K.~G. Murty and S.~N. Kabadi, ``Some np-complete problems in quadratic and nonlinear programming,'' \emph{Math. Program.}, vol.~39, no.~2, pp. 117--129, 1987.

\bibitem{bellare1995complexity}
M.~Bellare and P.~Rogaway, ``The complexity of approximating a nonlinear program,'' \emph{Math. Program.}, vol.~69, pp. 429--441, 1995.

\bibitem{pardalos1991quadratic}
P.~M. Pardalos and S.~A. Vavasis, ``Quadratic programming with one negative eigenvalue is np-hard,'' \emph{J. Global Optim.}, vol.~1, no.~1, pp. 15--22, 1991.

\bibitem{nesterov2008advance}
Y.~Nesterov, ``How to advance in structural convex optimization,'' \emph{OPTIMA: Mathematical Programming Society Newsletter}, vol.~78, pp. 2--5, 2008.

\bibitem{Qi2025}
Q.~Binchuan, G.~Wei, and L.~Li, ``Towards understanding the optimization mechanisms in deep learning,'' \emph{Applied Intelligence}, vol.~55, no.~15, p. 976, 2025.

\bibitem{vapnik1998statistical}
V.~Vapnik, ``Statistical learning theory,'' \emph{John Wiley \& Sons google schola}, vol.~2, pp. 831--842, 1998.

\bibitem{Bartlett2003RademacherAG}
P.~L. Bartlett and S.~Mendelson, ``Rademacher and gaussian complexities: Risk bounds and structural results,'' \emph{J. Mach. Learn. Res.}, vol.~3, pp. 463--482, 2003.

\bibitem{Yun2018SmallNI}
C.~Yun, S.~Sra, and A.~Jadbabaie, ``Small nonlinearities in activation functions create bad local minima in neural networks,'' in \emph{Proc. Int. Conf. Learn. Represent.}, 2019.

\bibitem{Arjevani2022AnnihilationOS}
Y.~Arjevani and M.~Field, ``Annihilation of spurious minima in two-layer relu networks,'' in \emph{Proc. Adv. Neural Inf. Process. Syst.}, vol.~35, 2022, pp. 37\,510--37\,523.

\bibitem{Jacot2018NeuralTK}
A.~Jacot, F.~Gabriel, and C.~Hongler, ``Neural tangent kernel: convergence and generalization in neural networks (invited paper),'' in \emph{{STOC}}.\hskip 1em plus 0.5em minus 0.4em\relax {ACM}, 2021, p.~6.

\bibitem{wang2023ntk}
Y.~Wang, D.~Li, and R.~Sun, ``Ntk-sap: Improving neural network pruning by aligning training dynamics,'' 2023, \textit{arXiv:2304.02840}.

\bibitem{liu2024ntk}
J.~Liu, Z.~Ji, Y.~Pang, and Y.~Yu, ``Ntk-guided few-shot class incremental learning,'' \emph{{IEEE} Trans. Image Process.}, vol.~33, pp. 6029--6044, 2024.

\bibitem{Sirignano2018MeanFA}
J.~A. Sirignano and K.~V. Spiliopoulos, ``Mean field analysis of neural networks: A central limit theorem,'' \emph{Stochastic Processes Appl.}, vol. 130, no.~3, pp. 1820--1852, 2018.

\bibitem{Mei2018AMF}
S.~Mei, A.~Montanari, and P.-M. Nguyen, ``A mean field view of the landscape of two-layer neural networks,'' \emph{Proc. Natl. Acad. Sci. U.S.A.}, vol. 115, pp. E7665 -- E7671, 2018.

\bibitem{kim2024transformers}
J.~Kim and T.~Suzuki, ``Transformers learn nonlinear features in context: Nonconvex mean-field dynamics on the attention landscape,'' in \emph{Proc. Int. Conf. Mach. Learn.}, 2024.

\bibitem{Seleznova2020AnalyzingFN}
M.~Seleznova and G.~Kutyniok, ``Analyzing finite neural networks: Can we trust neural tangent kernel theory?'' in \emph{Proc. Mach. Learn. Res.}, 2022, pp. 868--895.

\bibitem{Vyas2023EmpiricalLO}
N.~Vyas, Y.~Bansal, and P.~Nakkiran, ``Empirical limitations of the {NTK} for understanding scaling laws in deep learning,'' \emph{Trans. Mach. Learn. Res.}, vol. 2023, 2023.

\bibitem{Dauphin2014IdentifyingAA}
Y.~N. Dauphin, R.~Pascanu, {\c{C}}.~G{\"{u}}l{\c{c}}ehre, K.~Cho, S.~Ganguli, and Y.~Bengio, ``Identifying and attacking the saddle point problem in high-dimensional non-convex optimization,'' in \emph{Proc. Adv. Neural Inf. Process. Syst.}, 2014, pp. 2933--2941.

\bibitem{Keskar2016OnLT}
N.~S. Keskar, D.~Mudigere, J.~Nocedal, M.~Smelyanskiy, and P.~T.~P. Tang, ``On large-batch training for deep learning: Generalization gap and sharp minima,'' 2016, \textit{arXiv:1609.04836}.

\bibitem{Lee2019FirstorderMA}
J.~Lee, I.~Panageas, G.~Piliouras, M.~Simchowitz, M.~I. Jordan, and B.~Recht, ``First-order methods almost always avoid strict saddle points,'' \emph{Math. Program.}, vol. 176, pp. 311 -- 337, 2019.

\bibitem{zhou2023class}
Z.~Zhou, L.~Li, P.~Zhao, P.-A. Heng, and W.~Gong, ``Class-conditional sharpness-aware minimization for deep long-tailed recognition,'' in \emph{Proc. IEEE Conf. Comput. Vis. Pattern Recognit.}\hskip 1em plus 0.5em minus 0.4em\relax {IEEE}, 2023, pp. 3499--3509.

\bibitem{ding2024flat}
L.~Ding, D.~Drusvyatskiy, M.~Fazel, and Z.~Harchaoui, ``Flat minima generalize for low-rank matrix recovery,'' \emph{Information and Inference: A Journal of the IMA}, vol.~13, no.~2, p. iaae009, 2024.

\bibitem{le2024gradient}
B.~M. Le and S.~S. Woo, ``Gradient alignment for cross-domain face anti-spoofing,'' in \emph{Proc. IEEE Conf. Comput. Vis. Pattern Recognit.}\hskip 1em plus 0.5em minus 0.4em\relax {IEEE}, 2024, pp. 188--199.

\bibitem{zou2024flatten}
Y.~Zou, Y.~Liu, Y.~Hu, Y.~Li, and R.~Li, ``Flatten long-range loss landscapes for cross-domain few-shot learning,'' in \emph{Proc. IEEE Conf. Comput. Vis. Pattern Recognit.}, 2024, pp. 23\,575--23\,584.

\bibitem{dinh2017sharp}
L.~Dinh, R.~Pascanu, S.~Bengio, and Y.~Bengio, ``Sharp minima can generalize for deep nets,'' in \emph{Proc. Int. Conf. Mach. Learn.}, 2017, pp. 1019--1028.

\bibitem{ShwartzZiv2017OpeningTB}
R.~Shwartz-Ziv and N.~Tishby, ``Opening the black box of deep neural networks via information,'' 2017, \textit{arXiv:1703.00810}.

\bibitem{Ahuja2021InvariancePM}
K.~Ahuja, E.~Caballero, D.~Zhang, Y.~Bengio, I.~Mitliagkas, and I.~Rish, ``Invariance principle meets information bottleneck for out-of-distribution generalization,'' in \emph{Proc. Adv. Neural Inf. Process. Syst.}, 2021.

\bibitem{Wongso2023UsingSM}
S.~Wongso, R.~Ghosh, and M.~Motani, ``Using sliced mutual information to study memorization and generalization in deep neural networks,'' in \emph{Proc. Mach. Learn. Res.}, ser. Proceedings of Machine Learning Research, vol. 206, 2023, pp. 11\,608--11\,629.

\bibitem{saxe2019information}
A.~M. Saxe, Y.~Bansal, J.~Dapello, M.~Advani, A.~Kolchinsky, B.~D. Tracey, and D.~D. Cox, ``On the information bottleneck theory of deep learning,'' \emph{J. Stat. Mech: Theory Exp.}, vol. 2019, no.~12, p. 124020, 2019.

\bibitem{Oneto2023DoWR}
L.~Oneto, S.~Ridella, and D.~Anguita, ``Do we really need a new theory to understand over-parameterization?'' \emph{Neurocomputing}, vol. 543, p. 126227, 2023.

\bibitem{Hornik1989MultilayerFN}
K.~Hornik, M.~B. Stinchcombe, and H.~L. White, ``Multilayer feedforward networks are universal approximators,'' \emph{Neural Networks}, vol.~2, pp. 359--366, 1989.

\bibitem{Cybenko1989ApproximationBS}
G.~V. Cybenko, ``Approximation by superpositions of a sigmoidal function,'' \emph{Mathematics of Control, Signals and Systems}, vol.~2, pp. 303--314, 1989.

\bibitem{Leshno1993OriginalCM}
M.~Leshno, V.~Y. Lin, A.~Pinkus, and S.~Schocken, ``Original contribution: Multilayer feedforward networks with a nonpolynomial activation function can approximate any function,'' \emph{Neural Networks}, vol.~6, pp. 861--867, 1993.

\bibitem{Hanin2017ApproximatingCF}
B.~Hanin and M.~Sellke, ``Approximating continuous functions by relu nets of minimal width,'' 2017, \textit{arXiv:1710.11278}.

\bibitem{Kidger2019UniversalAW}
P.~Kidger and T.~Lyons, ``Universal approximation with deep narrow networks,'' in \emph{Proc. Mach. Learn. Res.}, 2020, pp. 2306--2327.

\bibitem{Zhang2023GoingDG}
J.~Zhang, T.~Liu, and D.~Tao, ``Going deeper, generalizing better: An information-theoretic view for deep learning.'' \emph{{IEEE} Trans. Neural Netw. Learn. Syst.}, vol.~35, no.~11, pp. 16\,683--16\,695, 2024.

\bibitem{Du2018GradientDP}
S.~S. Du, X.~Zhai, B.~P{\'{o}}czos, and A.~Singh, ``Gradient descent provably optimizes over-parameterized neural networks,'' in \emph{Proc. Int. Conf. Learn. Represent.}, 2019.

\bibitem{Poggio2019DoubleDI}
T.~A. Poggio, G.~Kur, and A.~Banburski, ``Double descent in the condition number,'' 2019, \textit{arXiv:1912.06190}.

\bibitem{Liu2021OnTD}
F.~Liu, J.~A.~K. Suykens, and V.~Cevher, ``On the double descent of random features models trained with {SGD},'' in \emph{NeurIPS}, vol.~35, 2022, pp. 34\,966--34\,980.

\bibitem{Hochreiter1997FlatM}
S.~Hochreiter and J.~Schmidhuber, ``Flat minima,'' \emph{Neural Comput.}, vol.~9, pp. 1--42, 1997.

\bibitem{Kaddour2022WhenDF}
J.~Kaddour, L.~Liu, R.~M.~A. Silva, and M.~J. Kusner, ``When do flat minima optimizers work?'' in \emph{Proc. Adv. Neural Inf. Process. Syst.}, vol.~35, 2022, pp. 16\,577--16\,595.

\bibitem{Wu2022TheAP}
L.~Wu, M.~Wang, and W.~J. Su, ``The alignment property of sgd noise and how it helps select flat minima: A stability analysis,'' in \emph{Proc. Adv. Neural Inf. Process. Syst.}, vol.~35, 2022, pp. 4680--4693.

\bibitem{Koehler2021UniformCO}
F.~Koehler, L.~Zhou, D.~J. Sutherland, and N.~Srebro, ``Uniform convergence of interpolators: Gaussian width, norm bounds, and benign overfitting,'' in \emph{Proc. Adv. Neural Inf. Process. Syst.}, vol.~34, 2021, pp. 20\,657--20\,668.

\bibitem{Chen2022UnderstandingBO}
L.~Chen, S.~Lu, and T.~Chen, ``Understanding benign overfitting in gradient-based meta learning,'' in \emph{Proc. Adv. Neural Inf. Process. Syst.}, vol.~35, 2022, pp. 19\,887--19\,899.

\bibitem{Vapnik2006EstimationOD}
V.~Vapnik, \emph{Estimation of Dependences Based on Empirical Data, Second Editiontion}.\hskip 1em plus 0.5em minus 0.4em\relax Springer, 2006.

\bibitem{Wu2021StatisticalLT}
Y.~Wu, ``Statistical learning theory,'' \emph{Technometrics}, vol.~41, pp. 377--378, 2021.

\bibitem{Bousquet2002StabilityAG}
O.~Bousquet and A.~Elisseeff, ``Stability and generalization,'' \emph{J. Mach. Learn. Res.}, vol.~2, pp. 499--526, 2002.

\bibitem{Oneto2015FullyEA}
L.~Oneto, A.~Ghio, S.~Ridella, and D.~Anguita, ``Fully empirical and data-dependent stability-based bounds,'' \emph{{IEEE} Trans. Cybern}, vol.~45, pp. 1913--1926, 2015.

\bibitem{ANDREASMAURER2017ASL}
A.~Maurer, ``A second-order look at stability and generalization,'' in \emph{Proc. Mach. Learn. Res.}, 2017, pp. 1461--1475.

\bibitem{ShaweTaylor1997APA}
J.~Shawe-Taylor and R.~C. Williamson, ``A pac analysis of a bayesian estimator,'' in \emph{Proc. Mach. Learn. Res.}, 1997, pp. 2--9.

\bibitem{McAllester1998SomePT}
D.~A. McAllester, ``Some pac-bayesian theorems,'' \emph{Machine Learning}, vol.~37, pp. 355--363, 1998.

\bibitem{Xie2017PACBayesBF}
X.~Xie and S.~Sun, ``Pac-bayes bounds for twin support vector machines,'' \emph{Neurocomputing}, vol. 234, pp. 137--143, 2017.

\bibitem{Alquier2021UserfriendlyIT}
P.~Alquier, ``User-friendly introduction to pac-bayes bounds,'' \emph{Found. Trends Mach. Learn.}, vol.~17, pp. 174--303, 2021.

\bibitem{Bottou2016OptimizationMF}
L.~Bottou, F.~E. Curtis, and J.~Nocedal, ``Optimization methods for large-scale machine learning,'' \emph{{SIAM} Rev.}, vol.~60, no.~2, pp. 223--311, 2018.

\bibitem{Jastrzebski2017ThreeFI}
S.~Jastrzebski, Z.~Kenton, D.~Arpit, N.~Ballas, A.~Fischer, Y.~Bengio, and A.~J. Storkey, ``Three factors influencing minima in sgd,'' 2018, \textit{arXiv:1711.04623}.

\bibitem{Lewkowycz2020TheLL}
A.~Lewkowycz, Y.~Bahri, E.~Dyer, J.~Sohl-Dickstein, and G.~Gur-Ari, ``The large learning rate phase of deep learning: the catapult mechanism,'' 2020, \textit{arXiv:2003.02218}.

\bibitem{Chizat2018OnTG}
L.~Chizat and F.~Bach, ``On the global convergence of gradient descent for over-parameterized models using optimal transport,'' in \emph{Proc. Adv. Neural Inf. Process. Syst.}, 2018, pp. 3040--3050.

\bibitem{Xu2017InformationtheoreticAO}
A.~Xu and M.~Raginsky, ``Information-theoretic analysis of generalization capability of learning algorithms,'' in \emph{Proc. Adv. Neural Inf. Process. Syst.}, vol.~30, 2017, pp. 2524--2533.

\bibitem{Hellstrm2022ANF}
F.~Hellstr{\"o}m and G.~Durisi, ``A new family of generalization bounds using samplewise evaluated cmi,'' in \emph{Proc. Adv. Neural Inf. Process. Syst.}, vol.~35, 2022, pp. 10\,108--10\,121.

\bibitem{Todd2003ConvexAA}
M.~J. Todd, \emph{Convex Analysis and Nonlinear Optimization: Theory and Examples}.\hskip 1em plus 0.5em minus 0.4em\relax Wiley Online Library, 2003.

\bibitem{10.1093/acprof:oso/9780199535255.001.0001}
S.~Boucheron, G.~Lugosi, and P.~Massart, \emph{Concentration Inequalities: A Nonasymptotic Theory of Independence}.\hskip 1em plus 0.5em minus 0.4em\relax Oxford University Press, 02 2013.

\bibitem{Beck2012SmoothingAF}
A.~Beck and M.~Teboulle, ``Smoothing and first order methods: A unified framework,'' \emph{SIAM J. Optim.}, vol.~22, pp. 557--580, 2012.

\bibitem{10.1214/aos/1176346255}
L.~Devroye, ``{The Equivalence of Weak, Strong and Complete Convergence in $L_1$ for Kernel Density Estimates},'' \emph{The Annals of Statistics}, vol.~11, no.~3, pp. 896 -- 904, 1983.

\bibitem{Blondel2019LearningWF}
M.~Blondel, A.~F. Martins, and V.~Niculae, ``Learning with fenchel-young losses,'' \emph{J. Mach. Learn. Res.}, vol.~21, no.~35, pp. 1--69, 2020.

\bibitem{Lin2003SomeEP}
G.-H. Lin, G.-H. Lin, and M.~Fukushima, ``Some exact penalty results for nonlinear programs and mathematical programs with equilibrium constraints,'' \emph{J. Optim. Theory Appl. (USA)}, vol. 118, pp. 67--80, 2003.

\bibitem{LeCun1998GradientbasedLA}
Y.~LeCun, L.~Bottou, Y.~Bengio, and P.~Haffner, ``Gradient-based learning applied to document recognition,'' \emph{Proc. {IEEE}}, vol.~86, no.~11, pp. 2278--2324, 1998.

\bibitem{DBLP:journals/corr/abs-1708-07747}
H.~Xiao, K.~Rasul, and R.~Vollgraf, ``Fashion-mnist: a novel image dataset for benchmarking machine learning algorithms,'' 2017, \textit{arXiv:1708.07747}.

\bibitem{krizhevsky2009learning}
A.~Krizhevsky and G.~Hinton, ``Learning multiple layers of features from tiny images,'' \emph{Handbook of Systemic Autoimmune Diseases}, vol.~1, no.~4, 2009.

\bibitem{DBLP:conf/cvpr/SzegedyLJSRAEVR15}
C.~Szegedy, W.~Liu, Y.~Jia, P.~Sermanet, S.~Reed, D.~Anguelov, D.~Erhan, V.~Vanhoucke, and A.~Rabinovich, ``Going deeper with convolutions,'' in \emph{Proc. Adv. Neural Inf. Process. Syst.}, 2015, pp. 1--9.

\bibitem{DBLP:conf/eccv/HeZRS16}
K.~He, X.~Zhang, S.~Ren, and J.~Sun, ``Identity mappings in deep residual networks,'' in \emph{Proc. Eur. Conf. Comput. Vis.}\hskip 1em plus 0.5em minus 0.4em\relax Springer, 2016, pp. 630--645.

\bibitem{DBLP:conf/cvpr/HuangLMW17}
G.~Huang, Z.~Liu, L.~Van Der~Maaten, and K.~Q. Weinberger, ``Densely connected convolutional networks,'' in \emph{Proc. IEEE Conf. Comput. Vis. Pattern Recognit.}, 2017, pp. 4700--4708.

\bibitem{DBLP:journals/corr/HowardZCKWWAA17}
A.~G. Howard, M.~Zhu, B.~Chen, D.~Kalenichenko, W.~Wang, T.~Weyand, M.~Andreetto, and H.~Adam, ``Mobilenets: Efficient convolutional neural networks for mobile vision applications,'' 2017, \textit{arXiv:1704.04861}.

\bibitem{DBLP:conf/eccv/MaZZS18}
N.~Ma, X.~Zhang, H.-T. Zheng, and J.~Sun, ``Shufflenet v2: Practical guidelines for efficient cnn architecture design,'' in \emph{Proc. Eur. Conf. Comput. Vis.}, 2018, pp. 116--131.

\bibitem{DBLP:journals/pami/HuSASW20}
J.~Hu, L.~Shen, S.~Albanie, G.~Sun, and E.~Wu, ``Squeeze-and-excitation networks,'' \emph{{IEEE} Trans. Pattern Anal. Mach. Intell.}, vol.~42, no.~8, pp. 2011--2023, 2020.

\bibitem{dosovitskiy2021an}
A.~Dosovitskiy, L.~Beyer, A.~Kolesnikov, D.~Weissenborn, X.~Zhai, T.~Unterthiner, M.~Dehghani, M.~Minderer, G.~Heigold, S.~Gelly \emph{et~al.}, ``An image is worth 16x16 words: Transformers for image recognition at scale,'' 2020, \textit{arXiv:2010.11929}.

\bibitem{DBLP:conf/cvpr/0002CHL025}
J.~Zhu, X.~Chen, K.~He, Y.~LeCun, and Z.~Liu, ``Transformers without normalization,'' in \emph{Proc. IEEE Conf. Comput. Vis. Pattern Recognit.}, 2025, pp. 14\,901--14\,911.

\bibitem{DBLP:conf/iccv/TouvronCSSJ21}
H.~Touvron, M.~Cord, A.~Sablayrolles, G.~Synnaeve, and H.~J{\'{e}}gou, ``Going deeper with image transformers,'' in \emph{Proc. IEEE Int. Conf. Comput. Vis.}, 2021, pp. 32--42.

\bibitem{yoshioka2024visiontransformers}
K.~Yoshioka, ``vision-transformers-cifar10: Training vision transformers (vit) and related models on cifar-10,'' \url{https://github.com/kentaroy47/vision-transformers-cifar10}, 2024.

\end{thebibliography}
\normalsize

\newpage
\appendices




\section*{Appendix for Proofs}

\label{appendix:proof}
In this section, we prove the results stated in the paper and provide necessary technical details and discussions.

\subsection{Proof of Proposition~\ref{prop:uniqueness_fenchel}}
\label{appendix:proof_uniqueness_fenchel} 
By setting $Q=\delta_{q}$, a Dirac delta distribution centered at $q$, it follows that $\mathbb{E}_{\omega \sim Q}[\ell(\omega, p)] = \ell(q, p)$. 
According to the expectation optimality condition, $\ell(q, p)$ achieves its global minimum value when $p = q$.
Since $\ell(q, p)$ is continuous, differentiable, and strictly convex with respect to $p$, it possesses a unique minimum at $p = q$. 
At this point, the gradient of $\ell(q, p)$ with respect to $p$ vanishes, i.e., $\nabla_{p} \ell(q, q) = 0$.
Thus, we may express the gradient of the loss function as:
\begin{equation*}
\nabla_{p}\ell(q,p) = (p - q)^\top g(q,p),
\end{equation*}
where $g(q,p)$ is a vector-valued function that captures the direction and rate of change of the loss with respect to $p$. Due to the uniqueness of the minimum value, the expected loss, $\mathbb{E}_{\omega \sim Q} [\ell(\omega, p)]$, is strictly convex, and at its minimum $\bar{\omega}$, where $\bar{\omega} = \mathbb{E}_{\omega \sim Q}[\omega]$, the gradient with respect to $p$ vanishes, i.e., $\nabla_p \mathbb{E}_{\omega\sim Q}[\ell(\omega,\bar{\omega})]=0$, which implies that $\sum_{\omega\in \mathcal{Q}} Q(\omega)(\bar{\omega}- \omega)^\top g(\omega,\bar{\omega})=0$, where $\omega\sim Q$. 

Given that $Q(\omega)$ is an arbitrary distribution function, $g(\omega,p)$ must be independent of $\omega$, allowing us to replace $g(\omega,p)$ with $g(p)$. 
Since $\mathbb{E}_{\omega\sim Q} [\ell(\omega, p)]$ is a strictly convex function, its Hessian with respect to $p$ can be derived as:
\begin{equation*}
\begin{aligned}
    \nabla^2_p \mathbb{E}_{\omega\sim Q} [\ell(\omega, p)]=\sum_{\omega\in \mathcal{Q}} Q(\omega)[g(p)+(p-\omega)^\top \nabla_p g(p)].
\end{aligned}    
\end{equation*}
Simplifying, we get: 
\begin{equation*}
        \nabla^2_p \mathbb{E}_{\omega\sim Q} [\ell(\omega, p)]=g(p)+(p-\bar{\omega})\nabla_p g(p)>0.
\end{equation*}
The inequality above is derived from the fact that $\nabla^2_p \mathbb{E}_{\omega\sim Q} [\ell(\omega, p)]$ is strictly convex.

When $p$ is set to $\bar{\omega}$, we have $\nabla^2_p \mathbb{E}_{\omega\sim Q} [\ell(\omega, p)] = g(\bar{\omega})>0$. 
Since $\bar{\omega}$ can take any value, it follows that for any arbitrary $p$, we get $g(p)> 0$. 

Let us define a strictly convex function $\Phi(p)$ with its Hessian matrix given by $\nabla^2_{p}\Phi(p)=g(p)$. 
The gradient of the loss function $\ell(q,p)$ with respect to $p$ can be expressed as follows:
\begin{equation*}
    \begin{aligned}
        \nabla_p \ell(q,p)&=\nabla^2_p\Phi(p)(p-q)=\nabla_p p_\Phi^*p-\nabla_p p_\Phi^*q\\
        &=\nabla_p(\langle p, p_\Phi^*\rangle -\Phi(p))-q^\top \nabla_p p_\Phi^*\\
        &=\nabla_p(\Phi^*(p_\Phi^*)-\langle q,p_\Phi^*\rangle +h(q)).
    \end{aligned}
\end{equation*}
Since $\nabla_p \ell(q,q)=0$, we find that $h(q)=\Phi^*(q_\Phi^*)-\langle q,q_\Phi^*\rangle+c=\Phi(q)+c$, where $c$ is a constant.
This leads to the following expression for the loss function:
\begin{equation*}
    \begin{aligned}
        &\ell(q,p)=\Phi^*(p_\Phi^*)-\langle q,p_\Phi^*\rangle +\Phi(q)+c,\ell(q,q)=c.
    \end{aligned}
\end{equation*}
It follows that:$\ell(q, p) - \ell(q, q) = d_\Phi(q, p_\Phi^*)$.

\subsection{Proof of Proposition~\ref{prop:decompose}}
\label{appendix:proof_decompose} 
Based on the definition of the Fenchel-Young loss, we have:\small
\begin{equation*}
    \begin{aligned}
       &\mathcal{L}_{\Phi}(q,f_\theta(\cdot)) = \mathbb{E}_{XY} \big[ \Phi(\mathbf{1}_Y) + \Phi^*(f_\theta(X)) - \mathbf{1}_Y^\top f_\theta(X) \big] \\
        &= \mathbb{E}_{X} \big[ \mathbb{E}_{Y \sim q_{\mathcal{Y}|X}} \Phi(\mathbf{1}_Y) + \Phi^*(f_\theta(X)) - q_{\mathcal{Y}|X}^\top f_\theta(X) \big] \\
        &=\mathrm{Ent}_\Phi(\mathbf{1}_Y|X) + \mathbb{E}_{X} \big[ d_\Phi(q_{\mathcal{Y}|X}, f_\theta(X)) \big].
    \end{aligned}
\end{equation*}
\normalsize
The conclusion for the empirical risk can be derived similarly.

\subsection{Proof of Lemma~\ref{lem:prop_thomo_fun}}
\label{appendix:proof_prop_thomo_fun}
According to the definition of the Legendre–Fenchel transform and applying the Cauchy–Schwarz inequality, we have\small
\begin{equation*}
    \begin{aligned}
        \Psi^*(\nu) 
        &= \max_{\mu} \left\{ \langle \mu, \nu \rangle - \Psi(\mu) \right\} \leq \max_{\mu} \{ \|\mu\|_2 \|\nu\|_2 - \frac{\|\mu a_{\Psi}\|_2^{r_\Psi}}{r_{\Psi}} \},
    \end{aligned}
\end{equation*}
\normalsize
where equality holds if and only if $\mu$ and $\nu$ are linearly dependent. Define $l(\mu) = \|\mu\|_2 \|\nu\|_2 - \frac{\|\mu a_{\Psi}\|_2^{r_\Psi}}{r_{\Psi}}$. Then, this function attains its maximum when$
    \|\mu\|_2 = \frac{ (\|\nu\|_2 r_{\Psi})^{\frac{1}{r_{\Psi}-1}} }{a_{\Psi}}$, and the corresponding maximum value is $
    \|\nu / a_{\Psi}\|_2^{\frac{r_{\Psi}}{r_{\Psi}-1}} \cdot \frac{r_{\Psi}-1}{r_{\Psi}}$.

Therefore, we obtain $
    \Psi^*(\nu) = \|\nu / a_{\Psi}\|_2^{\frac{r_{\Psi}}{r_{\Psi}-1}} \cdot \frac{r_{\Psi}-1}{r_{\Psi}}$.  Based on the definition of the norm power function, it follows that $
        r_{\Psi^*} = \frac{r_{\Psi}}{r_{\Psi} - 1}$ and $a_{\Psi^*} = a_{\Psi}$

Below, we proceed to prove Conclusion 2 
According to Euler's Homogeneous Function Theorem~\ref{lem:euler}, we have
\begin{equation*}
    \begin{aligned}
        r_\Psi\Psi(\mu)&=\mu^\top \nabla_{\mu} \Psi(\mu)=\mu^\top \mu_{\Psi}^*,\\
        r_{\Psi^*}\Psi^*(\mu^*_\Psi)&=(\mu_\Psi^*)^\top\nabla_{\mu^*_\Psi} \Psi^*(\mu_\Psi^*)=(\mu_{\Psi}^*)^\top \mu.
    \end{aligned}
\end{equation*}
Since $\Psi$ is strictly convex, it follows that $\mu^\top \mu_\Psi^* = \Psi(\mu) + \Psi^*(\mu_\Psi^*)$. Consequently, we have:
\begin{equation*}
    \begin{aligned}
        \Psi^*(\mu_\Psi^*) = (r_\Psi - 1)\Psi(\mu), \Psi(\mu) = (r_{\Psi^*} - 1)\Psi^*(\mu_\Psi^*).
    \end{aligned}
\end{equation*}
\subsection{Proof of Lemma~\ref{lem:convex_smooth}}
\label{appendix:proof_convex_smooth}

\noindent \textit{Proof of Conclusion 1}. Since $\Phi$ is a twice-differentiable function with a positive definite Hessian matrix, by Taylor mean value theorem, for any $\mu, \nu$ in the domain of $\Phi$, there exists $t \in [0,1]$ such that
\begin{equation*}
    S_\Phi(\mu, \nu_\Phi^*) = \frac{1}{2} (\mu - \nu)^\top \nabla^2 \Phi(\xi) (\mu - \nu),
\end{equation*}
where $\xi = t\mu + (1-t)\nu$. Since $\nabla^2 \Phi(\xi)$ is symmetric and positive definite, we can apply the Courant–Fischer min-max theorem to obtain the eigenvalue bounds:
\begin{equation*}
\begin{aligned}
        \lambda_{\min}(\nabla^2 \Phi(\xi)) \|\mu - \nu\|_2^2 &\leq (\mu - \nu)^\top \nabla^2 \Phi(\xi) (\mu - \nu) \\
        &\leq \lambda_{\max}(\nabla^2 \Phi(\xi)) \|\mu - \nu\|_2^2.
\end{aligned}
\end{equation*}
\noindent \textit{Proof of Conclusion 2}. Since $G(\mu) = d_\Phi(\mu,s)$. By the definition of the Fenchel-Young loss, we have:
\begin{equation*}
\begin{aligned}
d_{G}(\mu, \nu_G^*) &= G(\mu) + G^*(\nu_{F_y}^*) - \langle \mu, \nu_G^* \rangle \\
&= G(\mu) - G(\nu) - \langle \nu_G^*, (\mu - \nu) \rangle.
\end{aligned}
\end{equation*}
Substituting $G(\mu) = d_\Phi(\mu,s)$, we get $S_{G}(\mu, \nu_{G}^*) = S_\Phi(\mu, \nu_\Phi^*)$.
Since $\Phi$ is a strictly convex function, it follows that $d_\Phi(s,\mu) = d_{\Phi^*}(\mu,s)$.  
Given that $G(\mu) = d_\Phi(s,\mu)$, we obtain $G(\mu) = d_{\Phi^*}(\mu,s)$ and $S_G(\mu,\nu_G^*) = S_{\Phi^*}(\mu,\nu_{\Phi^*}^*)$.
Therefore, $d_\Phi(\mu, s)$ has the same extended convexity and smoothness properties as $\Phi(\mu)$, and $d_\Phi(s, \mu)$ has the same extended convexity and smoothness properties as $\Phi^*(\mu)$.

\noindent \textit{Proof of Conclusion 3}. We prove Conclusion 3 by construction. For notational convenience, define $G_{xy}(\theta) = d_\Phi(y, f_\theta(x))$. In practical neural network models, both the value and the gradient of $G_{xy}(\theta)$ are bounded within their domains.
Therefore, we can construct $\xi(\theta) = \|a_\xi \theta\|_2^{r_\xi} $, where $a_\xi \geq 0$. Clearly, in most cases, by appropriately increasing the parameters $a_\xi$ and $r_\xi$, we can ensure that for all $\theta_1, \theta_2 \in \Theta$ and $(x,y) \in (\mathcal{X},\mathcal{Y})$,
\begin{equation*}
    S_{G_{xy}}(\theta_1, \theta_2) \leq \|a_\xi(\theta_1 - \theta_2)\|_2^{r_\xi} .
\end{equation*}
Therefore, we can use $\xi(\cdot)$ to effectively bound $S_{G_{xy}}(\theta_1, \theta_2)$ for a wide range of functions $G_{xy}(\theta)$.

Since $\Phi$ is assumed to be strictly convex and differentiable, its conjugate $\Phi^*$ is also strictly convex and differentiable. Since $\Phi^*$ is differentiable and strictly convex, analogous to the construction of $\xi(\cdot)$, we can construct power functions $\Psi(\cdot)$ and $\psi(\cdot)$ such that for all $\theta \in \Theta$ and $(x,y) \in (\mathcal{X},\mathcal{Y})$,
\begin{equation*}
    \|a_\psi(f_\theta(x) - y_{\Phi}^*)\|_2^{r_\psi}  \leq S_{\Phi^*}(f_\theta(x), y_{\Phi}^*) \leq \|a_\Psi(f_\theta(x) - y_{\Phi}^*)\|_2^{r_\Psi} ,
\end{equation*}
where $y_{\Phi}^* = \nabla_y \Phi(y)$.
According to Conclusion 2, the extended convexity and smoothness of $d_\Phi(y, f_\theta(x))$ with respect to $f_\theta(x)$ are identical to those of $\Phi^*(f_\theta(x))$ with respect to $f_\theta(x)$. Therefore, the constructed functions $\Psi(\cdot)$ and $\psi(\cdot)$ guarantee that $d_\Phi(y, f_\theta(x))$ is $\mathcal{H}(\psi)$-convex and $\mathcal{H}(\Psi)$-smooth with respect to $f_\theta(x)$.

\subsection{Proof of Proposition~\ref{prop:h_bound}}
\label{appendix:proof_h_bound}

\noindent \textit{Proof of Conclusion 1}. According to the definition of $S_G(\nu,\mu)$, we have
\begin{equation*}\label{eq:def_base}
\begin{aligned}
        G_*&= \min_{\nu} G(\nu)= \min_{\nu} \{G(\mu)+(\mu_G^*)^\top (\nu-\mu) +S_G(\nu,\mu)\}\\
        &=\min_{\nu} \{G(\mu)+(\mu_G^*)^\top z+S_G(\nu,\mu)\},
\end{aligned}
\end{equation*} where $z=\nu-\mu$.
Since $ \mu_* \in \mathcal{G}_* $ represents a global minimum, which must also be a stationary point, it follows that $ \nabla G(\mu_*) = 0 $.
Since $G$ is $\mathcal{H}(\psi)$-smooth, we have 
\begin{equation*}
    \begin{aligned}
        G(\mu)&= G(\mu_*)+\nabla G(\mu_*)^\top (\mu-\mu_*) +S_G(\mu,\mu_*)\\
                 &\le G_*+\Psi(\mu-\mu_*).
    \end{aligned}
\end{equation*}
Thus, the upper bound of the first conclusion is proved.
Since $G$ is $\mathcal{H}(\Psi)$-smooth, we have $S_G(\nu,\mu)\le \Psi(z)$, where $z=\nu-\mu$.
By substituting the above inequality into formula~\eqref{eq:def_base}, it follows that 
\begin{equation*}
    G_*\le\min_z \{G(\mu)+ (\mu_G^*)^\top z+\Psi(z)\}.
\end{equation*} Letting $z_{\Psi}^*=-\mu_G^*$. Since $\Psi$ and $\Psi^*$ are even and homogeneous functions of degree $r_\Psi$ and $r_{\Psi^*}$, according to Lemma~\ref{lem:prop_thomo_fun:2}, we then have

\begin{equation*}
    \begin{aligned}
        r_{\Psi^*}\Psi^*(\mu_G^*)&=r_{\Psi^*}\Psi^*(z_{\Psi}^*)=( z_{\Psi}^*)^\top\nabla_{z_{\Psi}^*} \Psi^*(z_{\Psi}^*)\\
        &=(z_{\Psi}^*)^\top z=-z^\top \mu_G^*,\\
        r_\Psi\Psi(z)&=z^\top \nabla_z \Psi(z)=- (\mu_G^*)^\top z.
    \end{aligned}
\end{equation*}
Therefore, when $z_{\Psi}^*=-\mu_G^*$, we have 
\begin{equation*}\label{eq_nonconvex_1}
\begin{aligned}
   G_* &\le G(\mu)+ (\mu_G^*)^\top z- (\mu_G^*)^\top z/r_\Psi\\
    &=G(\mu)+ (\mu_G^*)^\top z/r_{\Psi^*}=G(\mu)-\Psi^*(\mu_G^*).
\end{aligned}
\end{equation*}
The lower bound of the first conclusion is thus proved.

\noindent \textit{Proof of Conclusion 2}. Since $ \mu_* \in \mathcal{G}_* $ represents a global minimum, which must also be a stationary point, it follows that $ \nabla G(\mu_*) = 0 $.
Since $G$ is $\mathcal{H}(\psi)$-convex, we have 
\begin{equation*}
    \begin{aligned}
        G(\mu)&=G(\mu_*)+\nabla G(\mu_*)^\top(\mu-\mu_*) +S_G(\mu,\mu_*)\\
                 &\ge G_*+\psi(\mu-\mu_*).
    \end{aligned}
\end{equation*}

Thus, the lower bound of the second conclusion is proved.
Since $G$ is $\mathcal{H}(\psi)$-convex, we have 
\begin{equation*}
\begin{aligned}
    G(\mu_*)&= G(\mu)+\nabla G(\mu)^\top(\mu_*-\mu) +S_G(\mu_*,\mu)\\
    &\ge G(\mu)+\nabla G(\mu)^\top(\mu_*-\mu) +\psi(\mu_*-\mu)\\
    &\ge \min_z \{G(\mu)+(\mu_G^*)^\top z+\psi(z)\},
\end{aligned}
\end{equation*}
where $z=\mu_*-\mu$.
When $z_{\psi}^*=-\mu_G^*$, the right side of the inequality attains its minimum value. 
Similarly, according to Euler's Homogeneous Function Theorem~\ref{lem:euler}, when $z_{\psi}^*=-\mu_G^*$, we have 
\begin{equation*}\label{eq_nonconvex_2}
    \begin{aligned}
     G_* &\ge G(\mu)+(\mu_G^*)^\top z-(\mu_G^*)^\top z/r_\psi\\
    &=G(\mu)+(\mu_G^*)^\top z/r_{\Psi^*}=G(\mu)-\psi^*(\mu_G^*).
    \end{aligned}
\end{equation*}

\noindent \textit{Proof of Conclusion 3}. Given that $G(\mu)$ is $\mathcal{H}(\psi)$-convex and $\mathcal{H}(\psi)$-smooth, by combining Equation~\eqref{eq_nonconvex_1} and Equation~\eqref{eq_nonconvex_2}, we obtain 
\begin{equation*}
    G(\mu)-\psi^*(\mu_G^*)\le G_*\le G(\mu)-\Psi^*(\mu_G^*).
\end{equation*}

\subsection{Proof of Proposition~\ref{prop:sgd_convergence}}
\label{appendix:proof_sgd_convergence} 

\noindent \textit{Proof of Conclusion 1 .}  Since $G(\theta,z)$ is $\mathcal{H}(\xi)$-smooth, by Proposition~\ref{prop:h_bound}, we have
\begin{equation*}
    G(\theta_{k+1},s_k) - G(\theta_k,s_k) + \alpha g(\theta_k,s_k)^\top g(\theta_k, s_k) \le \xi(\alpha g(\theta_k, s_k)).
\end{equation*}
It follows that
\begin{equation*}
\begin{aligned}
        G(\theta_{k+1},s_k) &- G(\theta_k,s_k)
        \le -\alpha \|g(\theta_k,s_k)\|_2^2 + \xi(\alpha g(\theta_k,s_k)) \\
        &= -\alpha \|g(\theta_k,s_k)\|_2^2 + \alpha^{r_\xi} \xi(g(\theta_k,s_k)) = K(\alpha),
\end{aligned}
\end{equation*}
where $K(\alpha) = -\alpha \|g(\theta_k,s_k)\|_2^2 + \alpha^{r_\xi} \xi(g(\theta_k,s_k))$.

$K(\alpha)$ attains its minimum at $ \alpha = \left( \frac{\|g(\theta_k,s_k)\|_2^2}{r_\xi \xi(g(\theta_k,s_k))} \right)^{\frac{1}{r_\xi - 1}}$, and the minimum value is
\begin{equation*}
    \min_\alpha K(\alpha) = -r_{\xi^*}^{-1} \left( \|g(\theta_k,s_k)\|_2^2 \right)^{\frac{r_\xi}{ r_\xi - 1}} \left( r_\xi \xi(g(\theta_k,s_k)) \right)^{-\frac{1}{r_\xi - 1}}.
\end{equation*}
Furthermore, $K(\alpha) \leq 0$ for $0 < \alpha < \left( \frac{\|g(\theta_k,s_k)\|_2^2}{\xi(g(\theta_k,s_k))} \right)^{1/(r_\xi - 1)}$.

Substituting the optimal $\alpha$, we obtain
\small
\begin{equation*}
    \begin{aligned}
        G(\theta_{k+1},s_k) &\le G(\theta_k,s_k) - r_{\xi^*}^{-1} \left( {\|g(\theta_k,s_k)\|_2^2}/{r_\xi \xi(g(\theta_k,s_k))} \right)^{r_{\xi^*}} \\
        &= G(\theta_k,s_k) - r_{\xi^*}^{-1} \left( {\|g(\theta_k,s_k)\|_2^2}/{\bar{\xi}(g(\theta_k,s_k))} \right)^{r_{\xi^*}}.
    \end{aligned}
\end{equation*}
\normalsize 

\noindent \textit{Proof of Conclusion 2.} Define $D(g(\theta_k,s_k)) = r_{\xi^*}^{-1} \left( {\|g(\theta_k,s_k)\|_2^2}/{\bar{\xi}(g(\theta_k,s_k))} \right)^{r_{\xi^*}}$. By assumption, we have
\small
\begin{equation*}
\begin{aligned}
        G(\theta_{k+1},s) - G(\theta_k,s) &\leq \beta \big( G(\theta_{k+1},s_k) - G(\theta_k,s_k) \big) \\
        &\leq -\beta D(g(\theta_k,s_k)).
\end{aligned}
\end{equation*}
\normalsize Summing over $k = 0, \ldots, T-1$, we obtain
\begin{equation*}
    G(\theta_T,s) \leq G(\theta_0,s) - \beta \sum_{k=0}^{T-1} D(g(\theta_k,s_k)).
\end{equation*}
Rearranging and using $G(\theta_T,s) \geq G_*$, we have $\sum_{k=0}^{T-1} D(g(\theta_k,s_k)) \leq {G(\theta_0,s) - G_*}/{\beta}$.
It follows that $
    \min_{k=0,\ldots,T-1} D(g(\theta_k,s_k)) \leq {G(\theta_0,s) - G_*}/{\beta T}$.
Thus, $D(g(\theta_k,s_k))$ is equivalent to $\|g(\theta_k,s_k)\|_2^{r_{\xi^*}}$ up to a constant factor. Specifically, there exists $c > 0$ such that $D(g(\theta_k,s_k)) \geq c \|g(\theta_k,s_k)\|_2^{r_{\xi^*}}$.

Taking expectations and combining the above, we get
\begin{equation*}
    \mathbb{E} \left[ \min_{k=0,\ldots,T-1} \|g(\theta_k,s_k)\|_2^{r_{\xi^*}} \right] \leq \frac{1}{c} \cdot \frac{G(\theta_0,s) - G_*}{\beta T}.
\end{equation*}
To ensure $\mathbb{E} \|g(\theta_k,s_k)\|_2^{r_{\xi^*}} \leq \varepsilon^{r_{\xi^*}}$, it suffices to set the right-hand side less than or equal to $\varepsilon^{r_{\xi^*}}$, which yields $T = \mathcal{O}(\varepsilon^{-r_{\xi^*}})$, completing the proof.

\subsection{Proof of Proposition~\ref{prop:st_H_eigenvalue_bound}}
\label{appendix:proof_st_H_eigenvalue_bound}

\noindent \textit{Proof of Conclution 1.}  By applying the chain rule for differentiation, we have
\begin{equation*}
            \|\nabla_\theta d_\Phi(\mathbf{1}_y,f_{\theta}(x))\|_2^2=e^\top A_xe,
\end{equation*}
where $A_x:=\nabla_\theta f_{\theta}(x)^\top \nabla_\theta f_{\theta}(x)$, $e:=\nabla_{f_{\theta}(z)} d_\Phi(\mathbf{1}_y, f_\theta(x))$.
Since $A_x$ is a symmetric matrix, we can apply the Courant–Fischer min-max theorem to derive the following inequality: 
\begin{equation*}
\begin{aligned}
    \lambda_{\min}(A_x) \|\nabla_{f_{\theta}(x)} &d_\Phi(\mathbf{1}_y, f_\theta(x))\|_2^2 \leq \|\nabla_\theta d_\Phi(\mathbf{1}_y, f_\theta(x))\|_2^2 \\
       &\leq \lambda_{\max}(A_x) \|\nabla_{f_{\theta}(x)} d_\Phi(\mathbf{1}_y, f_\theta(x))\|_2^2.
\end{aligned}
\end{equation*}
Since $\lambda_{\min}(A_x)\neq 0$ and $\|\mathbf{1}_y-f_\theta(x)_{\Phi^*}^*\|_2^2=\|\nabla_{f_\theta(x)} d_\Phi(\mathbf{1}_y, f_\theta(x))\|_2^2$, it follows that \small
\begin{equation*}\label{eq:l2_div}
\begin{aligned}
       {\|g(x,y)\|_2^2}/{ \lambda_{\max}(A_x)}&\le \|\mathbf{1}_y-f_\theta(x)_{\Phi^*}^*\|_2^2\le{\|g(x,y)\|_2^2}/{ \lambda_{\min}(A_x)}.
\end{aligned}
\end{equation*}
\normalsize Therefore, we have
\begin{equation*}\label{eq:uu_ll}
    \begin{aligned}
        \psi^*(\nabla_{f_\theta(x)}d_\Phi(\mathbf{1}_y, f_\theta(x))&\le {\psi^*(g(x,y))}/{\lambda_{\min}(A_x)^{r_{\psi^*}/2}}.
    \end{aligned}
\end{equation*}
According to Lemma~\ref{prop:h_bound}, we have \small
\begin{equation*} 
\begin{aligned}
d_\Phi(\mathbf{1}_y,f_{\theta}(x))  \leq \psi^*(\nabla_{f_\theta(x)}d_\Phi(\mathbf{1}_y, f_\theta(x)) .
\end{aligned}
\end{equation*}
\normalsize By taking the inequalities~\eqref{eq:uu_ll} into above inequality, we have\small
\begin{equation*}\label{eq:fy_loss}
\begin{aligned}d_\Phi(\mathbf{1}_y,f_{\theta}(x)) \leq {\psi^*(g(x,y))}/\lambda_{\min}(A_x)^{\frac{r_{\psi^*}}{2}} .
\end{aligned}
\end{equation*} \normalsize
\noindent \textit{Proof of Conclution 2.} Since $\lambda_{\min}(A_s) = \min_{x \in s} \lambda_{\min}(A_x)$, taking the expectation of inequality~\eqref{eq:l2_div} with respect to $(X,Y)$, we obtain
$2\mathrm{D}_{L_2^2/2}(Y,f_\theta(\cdot)_{\Phi^*}^*) \leq \mathbb{E}\left[{\|\nabla_\theta d_{\Phi}(\mathbf{1}_Y,f_\theta(X))\|_2^2}/{\lambda_{\min}(A_X)}\right]$.
Similarly, taking the expectation of inequality~\eqref{eq:fy_loss} with respect to $(X,Y)$, it follows that
\begin{equation*}
\begin{aligned}
    \mathcal{L}_\Phi(\hat{q},f_\theta(\cdot)) &- \min_\theta \mathcal{L}_\Phi(\hat{q},f_\theta(\cdot)) 
    = \mathbb{E}[d_\Phi(\mathbf{1}_Y,f_\theta(X))] \\
    &- \min_\theta\mathbb{E}[d_\Phi(\mathbf{1}_Y, f_\theta(X))] \\
    &\le\mathbb{E}[d_\Phi(\mathbf{1}_Y,f_\theta(X))]  - \mathbb{E}[\min_\theta d_\Phi(\mathbf{1}_Y, f_\theta(X))] \\
    &\le \frac{\mathbb{E}_{XY\sim \hat{q}} [\psi^*(\nabla_{\theta} d_\Phi(\mathbf{1}_Y, f_\theta(X)))]}{\lambda_{\min}(A_s)^{r_{\psi^*}/2}}.
\end{aligned}
\end{equation*}
\subsection{Proof of Proposition~\ref{prop:risk_bound}}
\label{appendix:proof_risk_bound} 
According to the definition of Fenchel-Young loss, we have
\begin{equation*}\label{eq:exp_d_phi}
\begin{aligned}
        &\min_\theta \mathbb{E}_X \left[d_\Phi(q_{\mathcal{Y}|X},f_{\theta}(X))\right] = \mathbb{E}_X \left[\Phi(q_{\mathcal{Y}|X})\right] \\
    &\quad\quad\quad\quad- \max_\theta \mathbb{E}_X \left[\langle q_{\mathcal{Y}|X},f_{\theta}(X)\rangle - \Phi^*(f_{\theta}(X))\right] \\
    &\le  \mathbb{E}_X \left[\Phi(q_{\mathcal{Y}|X})\right] - \mathbb{E}_X \left[\langle q_{\mathcal{Y}|X},(q_{\mathcal{Y}})_{\Phi}^*\rangle - \Phi^*((q_{\mathcal{Y}})_{\Phi}^*)\right] \\
    &= \mathbb{E}_X \left[\Phi(q_{\mathcal{Y}|X})\right] - \Phi(q_{\mathcal{Y}}) \\
    &= \mathrm{Ent}_\Phi(q_{\mathcal{Y}|X}),
\end{aligned}
\end{equation*}
where $(q_{\mathcal{Y}})_{\Phi}^*$ is the unique maximizer due to the strict convexity and differentiability of $\Phi$, and the last equality follows from the definition of generalized entropy.
By Proposition~\ref{prop:decompose}, the Fenchel-Young loss decomposes as:
\begin{equation*}
\begin{aligned}
\mathcal{L}_{\Phi}(q,f_\theta(\cdot)) &= \mathrm{Ent}_\Phi(\mathbf{1}_{Y}|X) + \mathbb{E}_{X\sim q_{\mathcal{X}}} d_\Phi(q_{\mathcal{Y}|X}, f_\theta(X)) \\
&\ge \mathrm{Ent}_\Phi(\mathbf{1}_{Y}|X),
\end{aligned}
\end{equation*}
since $d_\Phi(\cdot,\cdot) \geq 0$. Therefore, $
                0 \leq \min_{\theta} \mathcal{L}_{\Phi}(q,f_\theta(\cdot))-\mathrm{Ent}_\Phi(\mathbf{1}_{Y}|X) \leq \mathrm{Ent}_\Phi(q_{\mathcal{Y}|X})$.
Similarly, we obtain $ 0 \leq \min_{\theta} \mathcal{L}_{\Phi}(\hat{q},f_\theta(\cdot))-\mathrm{Ent}_\Phi(\mathbf{1}_Y|X) \leq \mathrm{Ent}_\Phi(\hat{q}_{\mathcal{Y}|X})$.
\subsection{Proof of Proposition~\ref{prop:prob_gen_bound}}
\label{appendix:proof_prob_gen_bound} 

Let $Z = f_*(X)$, and define the loss function $\ell(y, z) = d_\Phi(\mathbf{1}_y, z)$.  
By the definition of $\gamma = \max_{(x,y) \in \mathcal{X} \times \mathcal{Y}} d_\Phi(\mathbf{1}_y, f_*(x))$, it follows that, \small \begin{equation*}
\begin{aligned}
|\mathbb E_{\hat q} & \ell(Y,Z)-\mathbb E_{q} \ell(Y,Z)|=|\sum_{yz\in \mathcal{Y\times Z}} (\hat q(y,z)-q(y,z))\ell(y,z)| \\
&\le |\sum_{yz\in\mathcal{Y\times Z}}(q(y,z)-\hat q(y,z))(\ell(y,z)-\frac{\gamma}{2})|\\
&+|\sum_{yz\in\mathcal{Y\times Z}}\frac{\gamma}{2}(q(y,z)-\hat q(y,z))|\le \frac{\gamma}{2} \|q_{\mathcal{Z}\mathcal{Y}} - \hat{q}_{\mathcal{Z}\mathcal{Y}}\|_1,
\end{aligned}
\end{equation*}\normalsize
where $\hat{q}_{\mathcal{ZY}}$ denotes the empirical distribution of $(Z, Y)$ obtained from $n$ samples.

Now, applying Lemma~\ref{lem:l1_convergence}, which provides a concentration bound for the $\ell_1$-distance between true and empirical distributions, we have that when $ \frac{2\varepsilon}{\gamma} \ge \sqrt{\frac{20|\mathcal{Z}||\mathcal{Y}|}{n}} $, i.e., when $ \varepsilon \ge \gamma \sqrt{\frac{5(|\mathcal{X}|-\zeta)|\mathcal{Y}|}{n}} $ (noting that $|\mathcal{Z}| = |f_*(\mathcal{X})|$), it follows that \small
\begin{equation*}
    \begin{aligned}
        \Pr&\left( \left| \mathcal{L}_{\Phi}(q,f_*(\cdot)) - \mathcal{L}_{\Phi}(\hat{q},f_*(\cdot)) \right| \ge \varepsilon \right)\\
        & \le \Pr\left( \frac{\gamma}{2} \|q_{\mathcal{Z}\mathcal{Y}} - \hat{q}_{\mathcal{Z}\mathcal{Y}}\|_1 \ge \varepsilon \right) \\
        &= \Pr\left( \|q_{\mathcal{Z}\mathcal{Y}} - \hat{q}_{\mathcal{Z}\mathcal{Y}}\|_1 \ge \frac{2\varepsilon}{\gamma} \right) \le 3\exp\left(-\frac{4n\varepsilon^2}{25\gamma^2}\right).
    \end{aligned}
\end{equation*}
\normalsize
\subsection{Proof of Proposition~\ref{thm:regularization_generalization}}
\label{appendix:proof_regularization_generalization} 

Since $\theta = \mathbf{0}$ represents the minimum point of $\Phi^*(f_{\theta}(x))$, it follows that $\forall \theta \in \Theta'_\epsilon$:
\begin{equation*}
    \begin{aligned}
    &\nabla_{\theta}\Phi^*(f_{\mathbf{\theta}}(x))|_{\theta=\mathbf{0}} = \mathbf{0},\quad \nabla^2_{\theta} \Phi^*(f_{\theta}(x)) \succeq 0,\\
        &\lambda_{\max}(\nabla^2_{\theta} \Phi^*(f_{\theta}(x)))\ge\lambda_{\min}(\nabla^2_{\theta} \Phi^*(f_{\theta}(x)))\ge 0.
    \end{aligned}
\end{equation*}
By applying the mean value theorem, there exists $\theta' = \alpha \mathbf{0} + (1 - \alpha)\theta, \alpha \in [0, 1]$ such that the following inequalities hold:
\begin{equation*}
    \begin{aligned}
       \Phi^*(f_{\theta}(x)) &= \Phi^*(f_{\mathbf{0}}(x))+\theta^\top 
\nabla^2_{\theta}\Phi^*(f_{\theta'}(x))\theta.
        \end{aligned}
\end{equation*}
By applying the Courant–Fischer min-max theorem, we obtain the following inequality:
\begin{equation*}
\begin{aligned}
        \lambda_{\min}(\nabla^2_{\theta}\Phi^*(f_{\theta'}(x)))\|\theta\|_2^2 &\leq \Phi^*(f_{\theta}(x)) - \Phi^*(\mathbf{0}) \\
        &\leq \lambda_{\max}(\nabla^2_{\theta}\Phi^*(f_{\theta'}(x)))\|\theta\|_2^2.
\end{aligned}
\end{equation*}
Let $a = \min_{\theta \in \Theta'_\epsilon} \lambda_{\min}(\nabla^2_{\theta}\Phi^*(f_{\theta'}(x)))$ and $b = \max_{\theta \in \Theta'_\epsilon} \lambda_{\max}(\nabla^2_{\theta}\Phi^*(f_{\theta'}(x)))$. 
Then, for all $ \theta \in \Theta'_\epsilon $, we have:$a\|\theta\|_2^2 \leq \Phi^*(f_{\theta}(x)) - \Phi^*(\mathbf{0}) \leq b\|\theta\|_2^2$.
\subsection{Proof of Proposition~\ref{prop:generalize_explain}}
\label{appendix:proof_generalize_explain}

Let $Y'$ be a random variable taking values in $\mathcal{Y}$, with its conditional distribution given $X = x$ being uniform over $\mathcal{Y}$, i.e., $Y' | x \sim u$ for all $x \in \mathcal{X}$, where $u$ denotes the uniform distribution on $\mathcal{Y}$.
By the definition of the expected Fenchel–Young loss, we have:
\[
\begin{aligned}
    \mathbb{E}_{Y'} d_\Phi(\mathbf{1}_{Y'}, (f_\theta(x))_{\Phi^*}^*) &\leq \max_{y \in \mathcal{Y}} d_\Phi(\mathbf{1}_y, (f_\theta(x))_{\Phi^*}^*)\\
    &\leq |\mathcal{Y}| \cdot \mathbb{E}_{Y'} d_\Phi(\mathbf{1}_{Y'}, (f_\theta(x))_{\Phi^*}^*).
\end{aligned}
\]
Using Proposition~\ref{prop:decompose}, minimizing $\mathbb{E}_{Y'}d_\Phi(\mathbf{1}_{Y'}, (f_\theta(x))_{\Phi^*}^*)$ is equivalent to minimizing $d_\Phi(u, (f_\theta(x))_{\Phi^*}^*)$. 
From Proposition~\ref{thm:regularization_generalization}, there exist constants $a, b > 0$ and $\epsilon > 0$ such that for all $\theta \in \Theta'_\epsilon = \{\theta : \|\theta\|_2^2 \leq \epsilon\}$,
\[
a\|\theta\|_2^2 \leq d_{\Phi}\left((f_{\mathbf{0}}(x))_{\Phi^*}^*, f_\theta(x)\right) \leq b\|\theta\|_2^2.
\]
This implies that, locally around $\theta = \mathbf{0}$, minimizing $\|\theta\|_2^2$ is equivalent to minimizing $d_{\Phi}\left((f_{\mathbf{0}}(x))_{\Phi^*}^*, f_\theta(x)\right)$, and hence $d_\Phi(u, (f_\theta(x))_{\Phi^*}^*)$. As shown in the first part, minimizing this term reduces $\gamma = \max_{x,y} d_\Phi(\mathbf{1}_y, (f_\theta(x))_{\Phi^*}^*)$. Therefore, reducing $\|\theta\|_2^2$ for all $x \in \mathcal{X}$ (within a neighborhood of $\mathbf{0}$) is equivalent to reducing $\gamma$, completing the proof.


\section*{Biography Section}

\vfill

\end{document}